\documentclass[compsoc,conference,a4paper,10pt,times]{IEEEtran}
\IEEEoverridecommandlockouts
\usepackage{cite}
\usepackage{amsmath,amssymb,amsfonts}
\usepackage{algorithmic}
\usepackage{graphicx}
\usepackage{textcomp}
\usepackage{bmpsize}
\usepackage{xcolor}
\usepackage{lipsum}
\usepackage[colorlinks=true,urlcolor=black]{hyperref}
\def\BibTeX{{\rm B\kern-.05em{\sc i\kern-.025em b}\kern-.08em
    T\kern-.1667em\lower.7ex\hbox{E}\kern-.125emX}}

\usepackage{booktabs}   
\usepackage{titletoc}
\hypersetup{
    colorlinks = true,
    linkcolor = red,
    citecolor = blue,
    urlcolor = black
}
\usepackage{wrapfig}
\usepackage{tikz}
\usepackage{amsmath}
\usepackage{soul}
\usepackage{graphicx}
\usepackage[export]{adjustbox}
\usepackage{multirow}
\usepackage{tcolorbox}
\usepackage{enumitem}
 
\usepackage{colortbl}
\usepackage{pifont}
\newcommand{\cmark}{\text{\ding{51}}}
\newcommand{\xmark}{\text{\ding{55}}}
\usepackage{tabularx, ragged2e}
\newcolumntype{Y}{>{\centering\arraybackslash}X}
\usepackage{makecell}
\usepackage{CJKutf8} 
\usepackage[russian,english]{babel} 
\usepackage{hhline}

\definecolor{redcolor}{HTML}{ff4155}
\definecolor{aquacolor}{HTML}{7befe4}
\definecolor{orangecolor}{HTML}{fbbd5e}
\definecolor{pinkcolor}{HTML}{ff83b4}
\definecolor{coralcolor}{HTML}{ef6561}
\newcommand{\redcapsule}{\tikz[baseline=(c.base)]{
    \node[draw, rounded corners, inner sep=1pt,fill=redcolor,text=redcolor, line width=0.5pt] (c) {text};
}}
\newcommand{\aquacapsule}{\tikz[baseline=(c.base)]{
    \node[draw, rounded corners, inner sep=1pt,fill=aquacolor,text=aquacolor, line width=0.5pt] (c) {text};
}}
\newcommand{\orangecapsule}{\tikz[baseline=(c.base)]{
    \node[draw, rounded corners, inner sep=1pt,fill=orangecolor,text=orangecolor, line width=0.5pt] (c) {text};
}}
\newcommand{\pinkcapsule}{\tikz[baseline=(c.base)]{
    \node[draw, rounded corners, inner sep=1pt,fill=pinkcolor,text=pinkcolor, line width=0.5pt] (c) {text};
}}
\newcommand{\coralcapsule}{\tikz[baseline=(c.base)]{
    \node[draw, rounded corners, inner sep=1pt,fill=coralcolor,text=coralcolor, line width=0.5pt] (c) {text};
}}

\newcommand{\ballnumber}[1]{\tikz[baseline=(myanchor.base)] \node[circle,fill=.,inner sep=1pt] (myanchor) {\color{-.}\bfseries#1};}

\newcommand{\redballnumbersmall}[1]{\tikz[baseline=(c.base)]{
    \node[draw, rounded corners, inner sep=2pt, fill=redcolor, text=black] (c) {\footnotesize#1};
}}
\newcommand{\aquaballnumbersmall}[1]{\tikz[baseline=(c.base)]{
    \node[draw, rounded corners, inner sep=2pt, fill=aquacolor, text=black] (c) {\footnotesize#1};
}}
\newcommand{\orangeballnumbersmall}[1]{\tikz[baseline=(c.base)]{
    \node[draw, rounded corners, inner sep=2pt, fill=orangecolor, text=black] (c) {\footnotesize#1};
}}
\newcommand{\pinkballnumbersmall}[1]{\tikz[baseline=(c.base)]{
    \node[draw, rounded corners, inner sep=2pt, fill=pinkcolor, text=black] (c) {\footnotesize#1};
}}
\newcommand{\coralballnumbersmall}[1]{\tikz[baseline=(c.base)]{
    \node[draw, rounded corners, inner sep=2pt, fill=coralcolor, text=black] (c) {\footnotesize#1};
}}

\begin{document}


\title{SoK: Systematization and Benchmarking of Deepfake Detectors \\in a Unified Framework} 

\author{
\IEEEauthorblockN{Binh M. Le}
\IEEEauthorblockA{
{Sungkyunkwan University, S. Korea} \\
bmle@g.skku.edu}
\and
\IEEEauthorblockN{Jiwon Kim}
\IEEEauthorblockA{
{Sungkyunkwan University, S. Korea} \\
merwl0@g.skku.edu}
\and
\IEEEauthorblockN{Simon S. Woo\textsuperscript{*}}
\IEEEauthorblockA{
{Sungkyunkwan University, S. Korea} \\
swoo@g.skku.edu}
\thanks{\textsuperscript{*}Corresponding author.}
\and
\IEEEauthorblockN{Kristen Moore}
\IEEEauthorblockA{
{CSIRO’s Data61, Australia} \\
kristen.moore@data61.csiro.au}
\and
\IEEEauthorblockN{Alsharif Abuadbba}
\IEEEauthorblockA{
{CSIRO’s Data61, Australia} \\
sharif.abuadbba@data61.csiro.au}
\and
\IEEEauthorblockN{Shahroz Tariq}
\IEEEauthorblockA{
{CSIRO’s Data61, Australia} \\
shahroz.tariq@data61.csiro.au}
}

\maketitle

\begin{abstract}
Deepfakes have rapidly emerged as a serious threat to {society} due to their ease of creation and dissemination, triggering the accelerated development of detection technologies. However, many existing detectors rely on lab-generated datasets for validation, which may not prepare them for novel, real-world deepfakes. This paper extensively reviews and analyzes state-of-the-art deepfake detectors, evaluating them against several critical criteria. These criteria categorize detectors into \textbf{4} high-level groups and \textbf{13} fine-grained sub-groups, aligned with a unified conceptual framework we propose. This classification offers practical insights into the factors affecting detector efficacy. We evaluate the generalizability of \textbf{16} leading detectors across comprehensive attack scenarios, including black-box, white-box, and gray-box settings. Our systematized analysis and experiments provide a deeper understanding of deepfake detectors and their generalizability, paving the way for future research and the development of more proactive defenses against deepfakes. 
\end{abstract}


\section{Introduction}
\label{sec:intro}

The widespread use of deep learning to create deepfakes has raised significant concerns due to its misuse in the generation of malicious content and their indistinguishability from authentic content ~\cite{fbi2022,WDC_Metaverse, li2022seeing}. The easy access to user-friendly, open-source deepfake tools \cite{DeepFaceLab, faceswap, fom2019} further compounds the issue, posing serious cybersecurity and societal threats, such as its impacts on Facial Liveness Verification (FLV) systems ~\cite{li2022seeing}. As a consequence, researchers are actively working to enhance deepfake detection methods and strengthen existing detection systems \cite{woo2023qad, feng2023avad, bai2023aunet} through various analytical approaches, including spatial \cite{nguyen2019capsule, woo2023qad, tariq2021clrnet}, frequency \cite{qian2020f3net, song2022cdnet, ble2022add}, and temporal \cite{wang2023altfreezing} analyses, as well as identifying underlying artifacts or fingerprints \cite{tan2023lgrad}. However, the diversity and sophistication of deepfake attacks necessitate the development of detectors that are robust against novel manipulations such as noise \cite{jiang2020deeperforensics, haliassos2021lips}, compression \cite{ble2022add, woo2023qad}, and most critically, to identify unseen deepfakes in the wild \cite{pu2021deepfake, zhao2021multi, shiohara2022sbis}. This need is further emphasized by the limitations of current training datasets, which can leave detectors vulnerable to performance degradation against unseen deepfake variants, potentially resulting in {performance worse than} a random guess~\cite{pu2021deepfake,WDC_Why}. 

While some recent studies have asserted the robust generalizability of their model against various types of deepfakes \cite{haliassos2021lips, shiohara2022sbis}, their work has predominantly relied on standard academic datasets~\cite{Rossler2019ICCV, Celeb_DF_cvpr20}. This narrow focus has resulted in a limited understanding of deepfake detectors, generation tools, and datasets, particularly regarding their real-world functionalities, characteristics, and performance. Consequently, there is a significant gap between the reported efficacy of detectors and their actual performance, highlighting the critical need for comprehensive and systematic evaluations against a broad spectrum of deepfake tools and real-world scenarios. Previous efforts to systematically categorize generation and detection methods have not provided comprehensive thorough evaluations \cite{mirsky2021creation, seow2022comprehensive} or detailed classification of deepfake creation tools and advanced detectors \cite{yan2023deepfakebench}.
By conducting extensive, systematic evaluations against a diverse range of deepfake generation methods and real-world examples, this study seeks to close the knowledge gap regarding the efficacy of deepfake detectors, tools, and datasets. To the best of our knowledge, \textit{this study marks the first comprehensive endeavor to systematically scrutinize the existing body of research on deepfake detection}, aiming to address three pivotal research questions:

{\footnotesize
\begin{tcolorbox}
[width=1\linewidth, center,  left=10pt, right=0pt, top=0pt, bottom=0pt,label=rq1,boxrule=0.5pt]
\begin{enumerate}[leftmargin=15pt,label=\textbf{RQ\arabic*:},start=1]
    \item \textsc{What factors influence facial deepfake detection?}
    \item \textsc{How well do leading detectors generalize in performance?}
    \item \textsc{How do identified factors impact detectors generalizability?}
\end{enumerate}
\end{tcolorbox}
}

To address \textbf{\textit{RQ1}}, we conducted a systematic review of the literature from 2019 to 2023, selecting 51 top deepfake detectors. Our analysis identified 18 key factors that are critical to the construction of deepfake detectors, with those factors spanning deepfake types, artifact types, input data representation methods, network architectures, and training and evaluation styles.  We developed a conceptual framework for categorizing detectors by these factors, thereby enhancing our understanding and systematic evaluation of deepfake detection nuances. To tackle \textbf{\textit{RQ2}}, we introduced a rigorous evaluation framework to asses the generalizability of leading detectors through a security lens using black-box, gray-box, and white-box evaluation settings. To facilitate this evaluation, we created the first ever ``white-box'' deepfake dataset through a controlled process where key aspects like the deepfake generation tool, source and destination videos are stabilized. This framework allowed us to evaluate 16 SoTA detectors, assessing their adaptability to various deepfake scenarios. Our proposed comprehensive approach provided a nuanced understanding of detectors, directly addressing \textbf{\textit{RQ3}} by examining the influence of identified influential factors on the generalizability of detectors.

\begin{figure}[!t]
  \centering
  \frame{\includegraphics[width=0.47\textwidth]{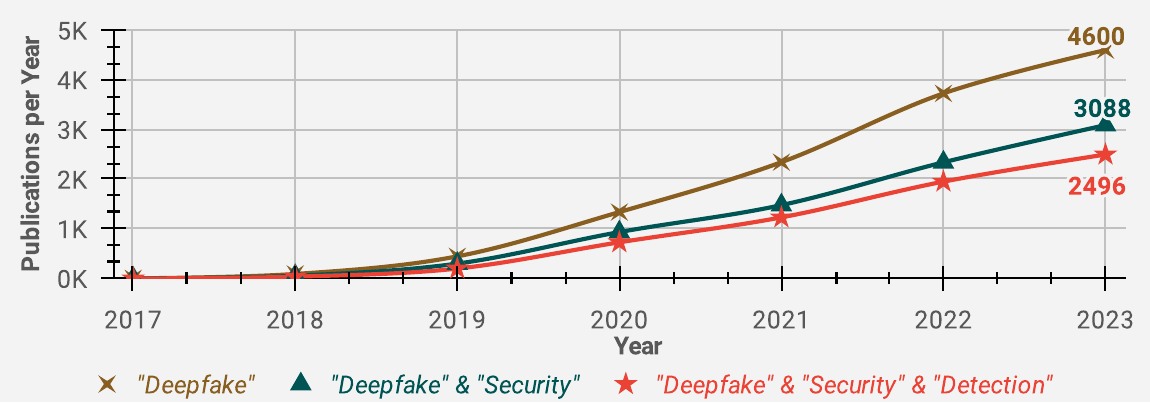}}
  \caption{Publications per year for deepfake-related keywords.}
  \label{fig:keyword_by_year}
\end{figure}

This study responds to the increase in deepfake-related publications by consolidating and systematizing the extensive body of existing research into a comprehensive analysis, as illustrated in Fig. \ref{fig:keyword_by_year} and detailed in Table \ref{tb:comparison_prior_work}. 
While numerous studies have explored deepfake detection and generation as shown in Fig.~\ref{fig:keyword_by_year}, \textit{a critical gap exists in research that systematically summarizes deepfake detectors under various influential factors and assesses their impact on well-known detectors using diverse protocol settings.}
In Table \ref{tab:detail_related_surveys}, we categorize and summarize prior survey studies across various criteria. 
The initial work by Verdoliva \cite{verdoliva2020media} focused primarily on deepfake detection but was too brief to provide in-depth insights into deepfake detectors and lacked detailed information on up-to-date detection methods.  Later, more thorough studies \cite{tolosana2020deepfakes, mirsky2021creation, juefei2022countering, seow2022comprehensive} provided comprehensive summaries covering various aspects of deepfake applications, threats, generation, and detection. Recent evaluation studies \cite{yan2023deepfakebench,khan2023deepfake,deng2024towards} have garnered more attention, yet they still lacked diversity in evaluation protocols, only considering gray-box settings. 
To the best of our knowledge, we are the first to present a thorough overview of the varying and dynamic deepfake detection landscape and to comprehensively evaluate it. Our paper distinguishes itself from previous surveys with the following unique features. (1) \textbf{Timeliness}, where we collect and analyze the latest SoTA deepfake detectors. (2) \textbf{Detail}, offering an analysis through an end-to-end conceptual framework and identifying influential factors; and  (3)  \textbf{Depth of Evaluation}, covering diverse settings: white-box, gray-box, and black-box, and providing more insights into detector performances through the lens of our framework.

\begin{table*}[h]
\centering
\caption{A detailed comparison of contributions between our SoK studies and relevant surveys. $\dagger$ indicates that the studies do not conduct experiments by themselves but solely report numbers.}
\label{tab:detail_related_surveys}
\resizebox{\linewidth}{!}{%
\begin{tabular}{l|c|c|c|c|c|cc|ccc|c} 
\hline\hline
\multirow{2}{*}{\textbf{Prior surveys}} & \multirow{2}{*}{\textbf{Year}} & \multirow{2}{*}{\begin{tabular}[c]{@{}c@{}}\textbf{Years Covered} \\ (\textit{Detectors})\end{tabular}} & \multirow{2}{*}{\begin{tabular}[c]{@{}c@{}}\textbf{Conceptual}\\\textbf{Framework}\end{tabular}} & \multirow{2}{*}{\begin{tabular}[c]{@{}c@{}}\textbf{Detectors}\\\textbf{Analysis}\end{tabular}} & \multirow{2}{*}{\begin{tabular}[c]{@{}c@{}}\textbf{Their Own}\\\textbf{Evaluation}\end{tabular}}  & \multicolumn{2}{c|}{\textbf{Evaluation Dataset}} & \multicolumn{3}{c|}{\textbf{(Cross) Evaluation Strategy}} & \multirow{2}{*}{\textbf{Notes}}\\ 
\cline{7-11}
 &  &  &  &  &  &  \textit{Same} & \textit{Cross} & \textit{Gray-box} & \textit{White-box} & \textit{Black-box} & \\ 
\hline\hline
Verdoliva~\cite{verdoliva2020media} & 2020 & $2005-2020$ & \textcolor{red}{\xmark} & Brief & \textcolor{red}{\xmark} &  \textcolor{green}{\cmark}$^\dagger$ & \textcolor{red}{\xmark} & \textcolor{red}{\xmark} & \textcolor{red}{\xmark} & \textcolor{red}{\xmark} &  Summarization \\ 
Tolosana et al.~\cite{tolosana2020deepfakes} & 2020 & $2018-2020$ & \textcolor{red}{\xmark} & Thorough & \textcolor{red}{\xmark}  & \textcolor{green}{\cmark}$^\dagger$ & \textcolor{red}{\xmark} & \textcolor{red}{\xmark} & \textcolor{red}{\xmark} & \textcolor{red}{\xmark} &  Summarization \\ 
Mirsky and Lee~\cite{mirsky2021creation} & 2021 & $2017-2020$ & \textcolor{red}{\xmark} & Thorough & \textcolor{red}{\xmark}  & \textcolor{green}{\cmark}$^\dagger$ & \textcolor{red}{\xmark} & \textcolor{red}{\xmark} & \textcolor{red}{\xmark} & \textcolor{red}{\xmark} &  Summarization\\ 
Juefei-Xu et al.~\cite{juefei2022countering} & 2022 & $2016-2021$ & \textcolor{red}{\xmark} & Thorough & \textcolor{red}{\xmark}  & \textcolor{green}{\cmark}$^\dagger$ & \textcolor{red}{\xmark} & \textcolor{red}{\xmark} & \textcolor{red}{\xmark} & \textcolor{red}{\xmark} &  Summarization \\ 
Rana et al.~\cite{rana2022deepfake} & 2022 & $2018-2020$ & \textcolor{red}{\xmark} & Brief & \textcolor{red}{\xmark}  & \textcolor{red}{\xmark} & \textcolor{red}{\xmark} & \textcolor{red}{\xmark} & \textcolor{red}{\xmark} & \textcolor{red}{\xmark} &  Summarization of Detectors \\ 
Nguyen et al.~\cite{nguyen2022deep} & 2022 & $2018-2021$ & \textcolor{red}{\xmark} & Brief & \textcolor{red}{\xmark}  & \textcolor{red}{\xmark} & \textcolor{red}{\xmark} & \textcolor{red}{\xmark} & \textcolor{red}{\xmark} & \textcolor{red}{\xmark} &  Summarization\\ 
Malik et al.~\cite{malik2022deepfake} & 2022 & $2018-2021$ & \textcolor{red}{\xmark} & Brief & \textcolor{red}{\xmark}  & \textcolor{red}{\xmark} & \textcolor{red}{\xmark} & \textcolor{red}{\xmark} & \textcolor{red}{\xmark} &  \textcolor{red}{\xmark} &  Summarization\\ 
Seow et al.~\cite{seow2022comprehensive} & 2022 & $2018-2021$ & \textcolor{red}{\xmark} & Thorough & \textcolor{red}{\xmark} & \textcolor{red}{\xmark} & \textcolor{red}{\xmark} & \textcolor{red}{\xmark} & \textcolor{red}{\xmark} &  \textcolor{red}{\xmark}&  Summarization\\ 
Naitali et al.~\cite{naitali2023deepfake} & 2023 & $2022-2023$ & \textcolor{red}{\xmark} & Brief & \textcolor{red}{\xmark} &  \textcolor{red}{\xmark} & \textcolor{red}{\xmark} & \textcolor{red}{\xmark} & \textcolor{red}{\xmark} & \textcolor{red}{\xmark} &  Summarization\\ 
Yan et al.~\cite{yan2023deepfakebench} & 2023 & $2018-2023$ & \textcolor{red}{\xmark} & Brief  & \textcolor{green}{\cmark} (15) & \textcolor{green}{\cmark} & \textcolor{green}{\cmark} & \textcolor{green}{\cmark} & \textcolor{red}{\xmark} & \textcolor{red}{\xmark} & \begin{tabular}[c]{@{}c@{}}Evaluation of Published Detectors\end{tabular}  \\ 
Khan \& Nguyen~\cite{khan2023deepfake} & 2023 & - & \textcolor{red}{\xmark} & Brief & \textcolor{green}{\cmark}  (8)  & \textcolor{green}{\cmark} & \textcolor{green}{\cmark} & \textcolor{green}{\cmark} & \textcolor{red}{\xmark} & \textcolor{red}{\xmark} & Evaluation of General NN Models\\ 
\hline
\textbf{Ours} & 2023 & $2019-2023$ & \textcolor{green}{\cmark} & Thorough & \textcolor{green}{\cmark} (16)  & \textcolor{green}{\cmark} & \textcolor{green}{\cmark} & \textcolor{green}{\cmark} & \textcolor{green}{\cmark} & \textcolor{green}{\cmark} & \begin{tabular}{c} 
Summarization \& Evaluation of \\
Published Detectors \end{tabular} \\
\hline\hline
\end{tabular}
}
\label{tb:comparison_prior_work}
\end{table*}

The primary contributions of our paper and their corresponding sections are as follows:

\indent \textbullet \, \textbf{Conceptual framework and influential factors: } We systematically review the recent literature, and introduce a conceptual framework for categorizing deepfake detectors based on 18 key factors essential to deepfake detection identified in \textbf{\textit{RQ1}} (Sec. \ref{sub:PaperSelectionCriteria},  \ref{sub:DetectorsAnalysis}, and \ref{sub:conceptual-framework}).\\
\indent \textbullet \, \textbf{Categorization and analysis of leading detectors:} We curate a list of 51 top detectors, and categorize them using our proposed framework (Sec. \ref{sub:taxonomy}).\\
\indent  \textbullet \, \textbf{Evaluation framework:} We develop a rigorous detector evaluation framework that includes black-box, gray-box, and white-box model evaluation settings, and create a novel white-box deepfake dataset. With these we perform a comprehensive assessment of 16 of the most recent SoTA detectors' performance, addressing \textbf{\textit{RQ2}} (Sec. \ref{sec:EvaluationSettings} and \ref{sec:experiments}). Our evaluation code is provided \href{https://anonymous.4open.science/r/deepfakeframework}{\textcolor{red}{\textit{here}}}.\\
\indent  \textbullet \, \textbf{Insights:} We explore the impact of identified influential factors on the generalizability of detectors, directly addressing \textbf{\textit{RQ3}} (Sec. \ref{sub:inpact_factors}).\\
\indent   \textbullet \, \textbf{Future Directions:} We use our framework to identify significant challenges facing current deepfake detection systems, as well as future pathways for enhancing deepfake detection (Sec. \ref{sec:future_direction}).

\section{Background and Related Work}
\label{sec:background}
\textbf{Deepfake Generation.}
The advent of Generative Adversarial Networks (GANs) by Goodfellow \textit{et al.}~\cite{goodfellow2020generative}, has advanced realistic image synthesis, especially for human faces \cite{stargan,progressivegan}. GANs use a generator ($\mathcal{G}$) and a discriminator ($\mathcal{D}$), training them adversarially. AutoEncoders (AE), initially proposed by LeCun \textit{et al.} \cite{lecun1987phd} and later refined as variational auto-encoders (VAEs) \cite{kingma2013auto}, compress data for altering face features in deepfake technology. The rise of deepfakes has spurred academic efforts like the Deepfake Detection Challenge (DFDC) \cite{dolhansky2020deepfake}, FaceForensics++ (FF++) \cite{Rossler2019ICCV}, Celebrity Deepfake (CelebDF) \cite{Celeb_DF_cvpr20}, and Audio-Video Deepfake (FakeAVCeleb) \cite{khalid2021fakeavceleb,HasamACMMM}. These deepfakes are generally categorized into face swaps, reenactments, and synthesis. 
For the reader's convenience, we denote the terms reference (source or driver) and target (destination or victim) identities as $\mathcal{R}$ and $\mathcal{T}$, respectively. $\mathcal{V}_\mathcal{R}$ signifies the reference video (perhaps sourced from the Internet), while $\mathcal{V}_\mathcal{T}$ refers to images or videos of the targeted individual. The deepfakes made from $\mathcal{V}_\mathcal{R}$ and $\mathcal{V}_\mathcal{T}$ are symbolized by $\mathcal{V}_\mathcal{D}$. 

\textit{Faceswap.}
Faceswap methods, such as FaceSwap~\cite{faceswap}, DeepFakes~\cite{deepfakes}, Faceshifter~\cite{li2019faceshifter}, and FSGAN~\cite{nirkin2019fsgan}, merge facial features from a target face ($\mathcal{V}_\mathcal{T}$) into a recipient video ($\mathcal{V}_\mathcal{R}$), creating a new video ($\mathcal{V}_\mathcal{D}$) where the target's face replaces the recipient's, while maintaining the original body and background. A notable example is superimposing a celebrity's face, like Scarlett Johansson's ($\mathcal{T}$), onto another person in a video ($\mathcal{V}_\mathcal{R}$)~\cite{ScarlettJohansson}, achieved using tools like DeepFaceLab~\cite{DeepFaceLab}, Dfaker~\cite{dfaker}, and SimSwap~\cite{Chen_2020_SimSwap}.

\textit{Reenactment.} 
The reenactment process combines $\mathcal{V}_\mathcal{T}$'s facial features with $\mathcal{V}_\mathcal{R}$'s expressions and movements to create $\mathcal{V}_\mathcal{D}$, using techniques like Talking Head (TH)~\cite{wang2021one}, First-Order Motion Model (FOM)~\cite{fom2019}, Face2Face~\cite{face2face}, and Neural Textures~\cite{neuraltexture}. This method has animated public figures, such as in altered speeches of Donald Trump~\cite{TrumpClimate} and Richard Nixon~\cite{NixonDisaster}.

\textit{Synthesis.} 
Synthesis deepfakes vary in method, with Diffusion or GAN recently surpassing facial blends in popularity. Focused on image synthesis ($\mathcal{I}$), these techniques blend identities, such as $\mathcal{R}^1$ and $\mathcal{R}^2$, to create a new image $\mathcal{I}_\mathcal{D}$. For instance, Diffusion can blend Donald Trump ($\mathcal{I}_\mathcal{R}^1$) and Joe Biden ($\mathcal{I}_\mathcal{R}^2$) to create a synthesized identity~\cite{TrumpDiffusion}.

\textbf{Deepfake Detection.}
Image forgery detection, especially concerning deepfakes with human faces, is extensively studied \cite{ForgeryDetectionSurvey}. Methods can be broadly categorized into supervised \cite{Rossler2019ICCV, zhao2021multi, tariq2021clrnet, MinhaCORED, CLRNetold, SamTAR, ShallowNet2, JeonghoPTD, SamGAN, MinhaFRETAL, ShallowNet1} and self-supervised approaches \cite{shiohara2022sbis, dong2022ict}. Self-supervised methods leverage large facial datasets to reduce bias but require hypothesizing artifact patterns. Supervised methods use deep learning to discern real from fake, often with varied input modalities \cite{li2020facexray, Rossler2019ICCV, cao2022recce, qian2020f3net, chen2021lrl, wang2023altfreezing, tariq2021clrnet, hu2022finfer}. Techniques target specific artifacts like mouth movements or gradients. Various detector categories exist, employing different architectures \cite{ble2022add, wang2023noisedf, bonettini2021b4att, coccomini2022ccvit, dong2022ict, feng2023avad, zheng2021ftcn}. However, a systematic evaluation of recent methods with unified criteria is lacking.

\section{Systematization of Deepfake Detectors}
\label{sec:method}

This section outlines our approach to select and assess deepfake detectors, targeting \textbf{\textit{RQ1}} (\textit{What factors influence facial deepfake detection?}) Utilizing insights from \textbf{51} deepfake studies, we developed a Conceptual Framework to categorize key concepts and relationships. Following rigorous selection criteria outlined in Sec.~\ref{sub:PaperSelectionCriteria}, we conducted a detailed review in Sec.~\ref{sub:DetectorsAnalysis}, focusing on aspects like dataset use, methodology, pre-processing, model architecture, and evaluation standards. This review informed the creation of our conceptual framework (Sec.~\ref{sub:conceptual-framework}), organizing detectors into 4 major groups and 13 detailed sub-groups (Sec.~\ref{sub:taxonomy}). This stage allows us to evaluate the most representative, open-source detectors from each group with standardized  metrics (Sec. \ref{sec:EvaluationSettings}), followed by influential factor assessment on those detectors (Sec.~\ref{sec:evaluation}).

\subsection{Paper Selection Criteria}
\label{sub:PaperSelectionCriteria}
First, we describe our paper collection process, including the inclusion and exclusion criteria.

\textbf{Paper Collection Process.} 
We focused on recent developments in deepfake detection in the last five years from 2019 to 2023, a period marked by significant growth in the field following the introduction of the FaceForensics++ benchmark~\cite{Rossler2019ICCV}. Utilizing the Google Scholar search query \texttt{``deepfake detection''} for this timeframe period, we identified 4,220 relevant publications.

\textbf{Inclusion and Exclusion Criteria.} 
We exclude papers not specifically related to deepfake detection and that do not propose a detector. Additionally, to ensure credibility, we exclude papers without a rigorous peer review, selecting only those published in CORE A* venues, except for some widely cited works, significantly reducing the pool\footnote{Note: We include two notable exceptions to our selection criteria: Capsule Forensics \cite{nguyen2019capsule} due to its high citation count 550+ and MCX-API \cite{xu2023mcx} due to its significant open source contributions and pretrained weights.}. Two authors independently reviewed the remaining papers and found that the majority works did not propose new detectors.  

This process yielded {51} relevant papers.  Note that many industry-developed detectors, like Intel FakeCatcher \cite{IntelFakeCatcher}, remain proprietary and closed-source, making them impractical to categorize or analyze within our frameworks. \textit{Next, we conducted a preliminary analysis of the 51 deepfake detectors. This analysis serves as the foundation for consolidating the deepfake detection pipeline into a conceptual framework, presented in the next section.}

\subsection{Preliminary Analysis of Detectors}
\label{sub:DetectorsAnalysis}
This section presents our analysis methodology for the 51 selected deepfake detectors, focusing initially on their primary detection targets—predominantly faceswap and reenactment deepfakes, with a minority (5 detectors) targeting synthetic image synthesis~\cite{sun2021lrnet, tan2023lgrad, qian2020f3net, wang2023noisedf, liu2021spsl}.

Moving on to artifact and pattern analysis, we observed that most detectors concentrate on spatial features independently, in conjunction with temporal or frequency domain features. Exceptions include two approaches~\cite{qian2020f3net, liu2021spsl}, where each exclusively focuses on the frequency domain, and three methods~\cite{wang2023noisedf, feng2023avad, fei2022sola}, which consider special artifacts such as Voice Sync and Noise Traces.

Our review of preprocessing techniques highlighted a variety of image processing, data augmentation, and face extraction methods, with detectors almost evenly split between single-frame and multi-frame data representations.

Exploring model architectures, we observed a dominance of deep neural networks, including ConvNets such as VGG \cite{simonyan2014very} and ResNet \cite{he2016deep}, sequence models like BiLSTM \cite{schuster1997bidirectional} and Vision Transformer \cite{dosovitskiy2020image}, in addition to specialized networks such as graph learning \cite{wang2023sfdg} or capsule networks \cite{nguyen2019capsule}. These DNNs were deployed in standalone configurations or in combination with each other, employing various learning strategies such as knowledge distillation \cite{hinton2015distilling}, Siamese networks \cite{bromley1993signature}. This investigation informed our understanding of architectural choices and artifact targeting across detectors.

Our investigation of validation methodologies revealed two main approaches: intra-dataset testing and cross-dataset testing to assess generalizability. Studies also adopted various evaluation metrics. 
This analytical endeavor yielded two key outcomes: (i) it elucidates the typical procedural steps followed by deepfake detectors, and (ii) it delineates the specific activities encompassed within these steps. This information serves as the foundation for consolidating the entire process into a Conceptual Framework.

\begin{figure*}[t]
    \centering
   \includegraphics[trim={19pt 19pt 19pt 20pt},clip, width=1\linewidth]{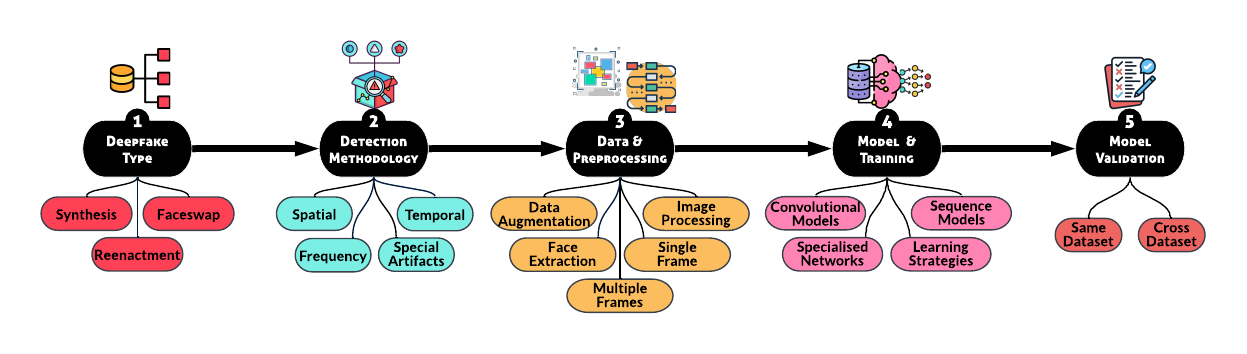}
    \caption{\textbf{Our Five-Step Conceptual Framework}: All detection methods adhere to this framework: Step \#1 (Deepfake Type), \#2 (Detection Methodology), \#3 (Data \& Preprocessing), \#4 (Model \& Training), and \#5 (Model Validation). From these primary stages, we identify 18 Influential Factors (illustrated as colored capsules) detailed in Sec. \ref{sub:conceptual-framework}.}
    \label{fig:ConceptualFramework}

\end{figure*}


\subsection{Conceptual Framework}
\label{sub:conceptual-framework}
Our review of the {51} selected papers on deepfake detection revealed a common five-step pipeline central to developing detection methods. This process forms the basis of our Conceptual Framework (\textbf{CF}), shown in Fig.~\ref{fig:ConceptualFramework}, featuring \textbf{18} Influential Factors (\textbf{IF}) (illustrated by \redcapsule{} \aquacapsule{} \orangecapsule{} \pinkcapsule{} \coralcapsule{} capsules) identified for \textbf{\textit{RQ1}}. Our \textbf{CF} components are described as follows, with \textbf{13 detailed sub-groups} provided in Sec.~\ref{sub:taxonomy}.

\noindent
{\ballnumber{1} \textbf{\textbf{Deeepfake Type.} }}
The first step of our framework involves identifying the specific type(s) of facial deepfake attacks that the detector will target. Fig. \ref{fig:ConceptualFramework} outlines the three categories of deepfakes considered in our framework, namely \redballnumbersmall{1A} \textit{Synthesis}, \redballnumbersmall{1B} \textit{Faceswap}, and \redballnumbersmall{1C} \textit{Reenactment}, which were mentioned in Sec. \ref{sec:background}. Recent literature on deepfake detectors primarily focus on faceswap and reenactment, as evidenced by \cite{mirsky2021creation}, \cite{masood2023deepfakes}.

\noindent
{\ballnumber{2} \textbf{Detection Methodology. }}
The second step involves detailing the detection methodology employed by detectors. These methodologies can be broadly classified into four main categories: \aquaballnumbersmall{2A} \textit{Spatial artifact}, \aquaballnumbersmall{2B} \textit{Temporal artifact}, \aquaballnumbersmall{2C} \textit{Frequency artifact}, and \aquaballnumbersmall{2D} \textit{Special artifact}-based detectors, each focusing on specific aspects of deepfake identification.

\textit{Spatial artifact}-based detectors analyze individual images or video frames for intra-frame visual anomalies like irregularities in texture, color, lighting, misalignments, or inconsistent blending between different segments of the image. 
\textit{Temporal artifact} detectors aim to identify inter-frame inconsistencies across multiple video frames over time. 

On the other hand, \textit{frequency artifact}-based detectors operate in the frequency domain. Deepfake manipulation often alters pixel value change rates, creating a distinctive frequency `signature' that affects the image's spectral characteristics, serving as discriminative cues for these detectors.

Additionally, \textit{special artifacts} focus on identifying unique manipulation signatures characteristic of deepfake generation methods. Examples include models detecting anomalies in synchronization features, such as audio-visual alignment between lip movement and voice \cite{feng2023avad}.

\noindent
{\ballnumber{3} \textbf{Data \& Preprocessing. }}
Our framework's third step focuses on the preparation and transformation of input data. We divide data preprocessing into three main areas: \orangeballnumbersmall{3A}~\textit{Data Augmentation}, \orangeballnumbersmall{3B}~\textit{Image Processing}, and \orangeballnumbersmall{3C}~\textit{Face Extraction}. Additionally, we classify its representation into two categories: \orangeballnumbersmall{3D}~\textit{Single-frame} and \orangeballnumbersmall{3E}~\textit{Multi-frame}.

\textit{Data Augmentation} plays the pivotal role of synthesizing training data, employing techniques such as Suspicious Forgeries Erasing \cite{wang2021rfm}, Self-Blended Images (SBI) \cite{shiohara2022sbis}, as well as Temporal Repeat and Dropout \cite{wang2023altfreezing}. Collectively, these methods strengthen the detector's ability to identify subtle anomalies indicative of deepfakes.

\textit{Image Preprocessing} techniques collectively contribute to the effective preparation and transformation of datasets, including methods such as 3D Dense Face Alignment (3DDFA) \cite{zhu2016face} to enable accurate feature extraction, and others such as Face Alignment \cite{bulat2017far} and RetinaFace with 4 key points \cite{deng2020retinaface} to ensure the standardization of facial features across images.

\textit{Face Extraction} techniques involve accurately identifying and isolating human faces in a video or image, using popular tools such as Dlib \cite{king2009dlib} and MTCNN \cite{zhang2016joint}.

\begin{table*}[!t]
\caption{\textbf{Systematic Classification of Deepfake Detectors.} 
In conceptual framework representations, white nodes indicate no papers fitting the category, half-colored nodes represent partial category representation, and fully colored nodes signify complete representation within the category (see Supp. Table \ref{app_tab:DetectorFurtherDetails} for details on detectors). The "FF++ Score" column displays each detector's performance on the FF++ dataset. Detectors marked with $\dagger$ were selected for further evaluations in Sec. \ref{sec:EvaluationSettings}. }
\label{tab:detectors-methodology}
\resizebox{1\linewidth}{!}{%
\begin{tabularx}{1.2\linewidth}{l|l|l|c|l|c|l|l}
\cline{1-8}\cline{1-8}
\multicolumn{2}{c|}{\thead{\textsc{\textbf{Focus of }}\\\textsc{\textbf{Methodology}}}}
& \thead{\textsc{\textbf{Distinct Technique of}}\\\textsc{\textbf{ Detector Architecture}}}
& \thead{\textsc{\textbf{Conceptual Framework}}\\\textsc{\textbf{ Representation}}}
& \thead{\textsc{\textbf{Venue}}}
& \thead{\textsc{\textbf{Year}}}
& \thead{\textsc{\textbf{Detector}}\\\textsc{\textbf{Name}}}
& \thead{\textsc{\textbf{FF++}}\\\textsc{\textbf{Score}}}\\
\cline{1-8}\cline{1-8}
\multirow{24}{*}{\rotatebox[origin=c]{90}{\textsc{\textbf{Spatial Artifacts}}}} 
    & \multirow{8}{*}{\begin{tabular}[r]{@{}r@{}}\textbf{CF \#1. }\textit{ConvNet}\\\textit{Models}\end{tabular}} & \textit{Capsule Network} & \multirow{8}{0.32\linewidth}{\includegraphics[trim={0pt 17pt 0pt 15pt},clip, width=1\linewidth,valign=b,page=2]{Figures/ConceptualFrameworkCategory.pdf}}
      & \textsc{icassp} & \textquotesingle19  &  Cap.Forensics$^\dagger$\cite{nguyen2019capsule} 
       & 96.60 \textsc{(auc)}  \\
     & & \textit{Depthwise Convolutions} & & \textsc{iccv} &\textquotesingle19 & XceptionNet$^\dagger$\cite{Rossler2019ICCV} & 99.26 \textsc{(acc)} \\
     & & \textit{Face X-ray Clues} & & \textsc{cvpr} & \textquotesingle20   & Face X-ray\cite{li2020facexray}& 98.52 \textsc{(auc)} \\
     & & \textit{Unified Methodology} &  & \textsc{cvpr} & \textquotesingle20  & FFD\cite{stehouwer2019ffd} & \multicolumn{1}{c}{-} \\
     & & \textit{Bipartite Graphs} & & \textsc{\textsc{cvpr}} & \textquotesingle22  & RECCE\cite{cao2022recce} & 99.32 \textsc{(auc)} \\
     & & \textit{Consistency Loss} & & \textsc{cvprw} & \textquotesingle22  & CORE\cite{ni2022core} & 99.94 \textsc{(auc)} \\
     & & \textit{Face Implicit Identities} & & \textsc{cvpr} & \textquotesingle23  & IID\cite{huang2023iid}  & 99.32 \textsc{(auc)} \\
     & & \textit{Multiple Color Spaces} &   & \textsc{wacvw} & \textquotesingle23  & MCX-API$^\dagger$\cite{xu2023mcx} & 99.68 \textsc{(auc)} \\ 
\cline{2-8}
    & \multirow{5}{*}{\begin{tabular}[r]{@{}r@{}} \textbf{CF \#2. } \textit{Specialized}\\\textit{Networks}\end{tabular}} & \textit{Siamese Training} &   \multirow{5}{0.32\linewidth}{\includegraphics[trim={0pt 17pt 0pt 16pt},clip, width=1\linewidth,valign=b,page=5]{Figures/ConceptualFrameworkCategory.pdf}}
      & \textsc{icpr} & \textquotesingle20  & EffB4Att$^\dagger$\cite{bonettini2021b4att} & 94.44 \textsc{(auc)} \\
     & & \textit{Intra-class Compact Loss} & & \textsc{aaai} & \textquotesingle21  & LTW\cite{sun2021ltw} & 99.17 \textsc{(auc)} \\
     & & \textit{Multi-attention losses} & & \textsc{cvpr} & \textquotesingle21 &MAT$^\dagger$\cite{zhao2021multi}  & 99.27 \textsc{(auc)} \\
     & & \textit{Intra-instance CL} && \textsc{aaai} & \textquotesingle22   & DCL\cite{sun2022dcl}  & 99.30 \textsc{(auc)} \\
     & & \textit{Self-blend Image} && \textsc{cvpr} & \textquotesingle22  &SBIs$^\dagger$\cite{shiohara2022sbis}  & 99.64 \textsc{(auc)} \\
\cline{2-8}
    & \multirow{5}{*}{\begin{tabular}[l]{@{}l@{}}\textbf{CF \#3. } \textit{ConvNet}\\\textit{Models with Learning }\\\textit{Strategies}\end{tabular}} & \multirow{5}{*}{\begin{tabular}[c]{@{}l@{}}\textit{Adversarial Learning} \\ \textit{High Frequency Pattern} \\ \textit{Meta-learning} \end{tabular}}  
    & \multirow{5}{0.32\linewidth}{\includegraphics[trim={0pt 16pt 0pt 15pt},clip, width=1\linewidth,valign=b,page=8]{Figures/ConceptualFrameworkCategory.pdf}}
    & \multirow{5}{*}{\begin{tabular}[c]{@{}l@{}}\textsc{acmmm} \\ \textsc{cvpr} \\ \textsc{neurips} \end{tabular}} 
    & \multirow{5}{*}{\begin{tabular}[c]{@{}l@{}}\textquotesingle21 \\ \textquotesingle21 \\ \textquotesingle22 \end{tabular}} 
    & \multirow{5}{*}{\begin{tabular}[c]{@{}l@{}}MLAC\cite{cao2021mlac} \\ FRDM\cite{luo2021frdm} \\ OST\cite{chen2022ost} \end{tabular}} 
    & \multirow{5}{*}{\begin{tabular}[c]{@{}l@{}}88.29 \textsc{(auc)} \\ \multicolumn{1}{c}{-} \\ 98.20 \textsc{(auc)} \end{tabular}} \\ 
    
     &&&&&&&\\ 
     &&&&&&&\\ 
     &&&&&&&\\ 
     &&&&&&&\\ 
\cline{2-8}
     & \multirow{5}{*}{\begin{tabular}[l]{@{}l@{}}\textbf{CF \#4.} \textit{ConvNet with}\\\textit{Specialized Networks}\end{tabular}}
     & \multirow{5}{*}{\begin{tabular}[c]{@{}l@{}}\textit{Identity Representation}\\\textit{Collaborative Learning} \end{tabular}}
     & \multirow{5}{0.32\linewidth}{\includegraphics[trim={0pt 16pt 0pt 10pt},clip, width=1\linewidth,valign=b,page=11]{Figures/ConceptualFrameworkCategory.pdf}}
     & \multirow{5}{*}{\begin{tabular}[c]{@{}l@{}}\textsc{cvpr}\\\textsc{iccv} \end{tabular}}
     & \multirow{5}{*}{\begin{tabular}[c]{@{}l@{}}\textquotesingle23\\\textquotesingle23 \end{tabular}}
     & \multirow{5}{*}{\begin{tabular}[c]{@{}l@{}}CADDM$^\dagger$\cite{dong2023caddm}\\
     QAD\cite{woo2023qad}\end{tabular}}  
     & \multirow{5}{*}{\begin{tabular}[c]{@{}l@{}}\textsc{99.70 \textsc{(auc)}}\\\textsc{95.60 \textsc{(auc)}}\end{tabular}}\\
      &&&&&&&\\
      &&&&&&&\\
      &&&&&&&\\ 
     &&&&&&&\\
\cline{2-8}
    & \multirow{5}{*}{\begin{tabular}[r]{@{}r@{}}\textbf{CF \#5.} \textit{Sequence}\\\textit{Models}\end{tabular}}
    & \multirow{5}{*}{\begin{tabular}[c]{@{}l@{}}\textit{Facial \& Other Inconsistency} \\ \textit{Unsupervised Inconsistency} \\ \textit{Action Units} \end{tabular}}  &  \multirow{5}{0.32\linewidth}{\includegraphics[trim={0pt 16pt 0pt 11pt},clip, width=1\linewidth,valign=b,page=14]{Figures/ConceptualFrameworkCategory.pdf}}
    & \multirow{5}{*}{\begin{tabular}[c]{@{}l@{}}\textsc{cvpr} \\ \textsc{eccv} \\ \textsc{cvpr} \end{tabular}} 
    & \multirow{5}{*}{\begin{tabular}[c]{@{}l@{}}\textquotesingle22 \\ \textquotesingle22 \\ \textquotesingle23 \end{tabular}} 
    & \multirow{5}{*}{\begin{tabular}[c]{@{}l@{}}ICT$^\dagger$\cite{dong2022ict} \\ UIA-ViT\cite{zhuang2022uiavit} \\ AUNet\cite{bai2023aunet} \end{tabular}} 
    & \multirow{5}{*}{\begin{tabular}[c]{@{}l@{}}98.56 \textsc{(auc)} \\ 99.33 \textsc{(auc)} \\ 99.89 \textsc{(auc)} \end{tabular}} \\ 
     &&&&&&&\\ 
     &&&&&&&\\
     &&&&&&&\\
     &&&&&&&\\

\cline{1-8}\cline{1-8}

\multirow{17}{*}{\rotatebox[origin=c]{90}{\textsc{\textbf{SpatioTemporal Artifacts}}}}
    & \multirow{9}{*}{\begin{tabular}[r]{@{}r@{}}\textbf{CF \#6.} \textit{ConvNet}\\\textit{Models}
    \end{tabular}} & \textit{Facial Attentive Mask} & \multirow{9}{0.32\linewidth}{\includegraphics[trim={0pt 16pt 0pt 14pt},clip, width=1\linewidth,valign=b,page=17]{Figures/ConceptualFrameworkCategory.pdf}}
     & \textsc{acmmm} & \textquotesingle20 & ADDNet-3d\cite{zi2020addnet3d}   & 86.69 \textsc{(acc)}  \\
     & & \textit{Anomaly Heartbeat Rhythm} & & \textsc{acmmm} & \textquotesingle20  & DeepRhythm\cite{qi2020deeprhythm}   & 98.50 \textsc{(acc)}  \\
     & & \textit{Multi-instance Learning} && \textsc{acmmm} & \textquotesingle20  & S-IML-T\cite{li2020simlt}   & 98.39 \textsc{(acc)}  \\
     & & \textit{Time Discrepancy Modeling} &  & \textsc{ijcai} & \textquotesingle21  & TD-3DCNN\cite{zhang2021td3dcnn} & 72.22 \textsc{(auc)} \\
     & &  \textit{Global-Local frame learning} & & \textsc{ijcai} & \textquotesingle21   & DIA\cite{hu2021dia}   & 98.80 \textsc{(auc)}  \\ 
     & & \textit{Local Dynamic Sync} &  & \textsc{aaai} & \textquotesingle22 & DIL\cite{gu2022dil}  & 98.93 \textsc{(acc)}  \\
     & & \textit{Faces Predictive Learning} && \textsc{aaai} & \textquotesingle22 & FInfer\cite{hu2022finfer}   & 95.67 \textsc{(auc)}  \\
     && \textit{Contrastive Learning} && \textsc{eccv} & \textquotesingle22  & HCIL\cite{gu2022hcil}   & 99.01 \textsc{(acc)}   \\
     & & \textit{Alternate Modules Freezing} & & \textsc{cvpr} & \textquotesingle23  & AltFreezing$^\dagger$\cite{wang2023altfreezing}  & 98.60 \textsc{(auc)}   \\
\cline{2-8}
    & \multirow{5}{*}{\begin{tabular}[l]{@{}l@{}}\textbf{CF \#7. }\textit{ConvNet with}\\\textit{Specialized Networks}\\\textit{\& Learning Strategies}\end{tabular}} 
    & \multirow{5}{*}{\begin{tabular}[c]{@{}l@{}}\textit{SpatioTemporal Inconsistency}\\\textit{Reading Mouth Movements}\\ \textit{Temporal Transformer}\end{tabular}}
     & \multirow{5}{0.32\linewidth}{\includegraphics[trim={0pt 16pt 0pt 15pt},clip, width=1\linewidth,valign=b,page=20]{Figures/ConceptualFrameworkCategory.pdf}}
     &  \multirow{5}{*}{\begin{tabular}[c]{@{}l@{}}\textsc{acmmm}\\ \textsc{cvpr}\\ \textsc{iccv}\end{tabular}} & \multirow{5}{*}{\begin{tabular}[c]{@{}l@{}}\textquotesingle21\\ \textquotesingle21\\ \textquotesingle21\end{tabular}}& \multirow{5}{*}{\begin{tabular}[c]{@{}l@{}}STIL\cite{gu2021stil}\\ LipForensics$^\dagger$\cite{haliassos2021lips}\\ FTCN$^\dagger$\cite{zheng2021ftcn}\end{tabular}}   &  \multirow{5}{*}{\begin{tabular}[c]{@{}l@{}}98.57 \textsc{(acc)}\\97.10 \textsc{(auc)}\\ \multicolumn{1}{c}{-}\end{tabular}}\\
     &&&&&&&\\ 
     &&&&&&&\\
     &&&&&&&\\ 
     &&&&&&&\\
\cline{2-8}

& \multirow{5}{*}{\begin{tabular}[r]{@{}r@{}}\textbf{CF \#8. }\textit{Sequence}\\\textit{Models}\end{tabular}} 

& \multirow{5}{*}{\begin{tabular}[c]{@{}l@{}}\textit{Combine ViT and CNN} \\ \textit{Spatial-temporal Modules} \\ \textit{Unsupervised Learning} \end{tabular}}  &  \multirow{5}{0.32\linewidth}{\includegraphics[trim={0pt 12pt 0pt 15pt},clip, width=1\linewidth,valign=b,page=23]{Figures/ConceptualFrameworkCategory.pdf}}
 &     \multirow{5}{*}{\begin{tabular}[c]{@{}l@{}}\textsc{iciap} \\ \textsc{www}  \\ \textsc{n}eur\textsc{ips} \end{tabular}} 
  & \multirow{5}{*}{\begin{tabular}[c]{@{}l@{}}\textquotesingle21  \\ \textquotesingle21  \\ \textquotesingle22 \end{tabular}}        & \multirow{5}{*}{\begin{tabular}[c]{@{}l@{}}CCViT$^\dagger$\cite{coccomini2022ccvit} \\ CLRNet$^\dagger$\cite{tariq2021clrnet} \\ LTTD\cite{guan2022lttd} \end{tabular}} 
  & \multirow{5}{*}{\begin{tabular}[c]{@{}l@{}}80.00 \textsc{(acc)} \\ 99.35 \textsc{(f$_1$)} \\ 97.72 \textsc{(auc)} \end{tabular}}  \\
 & &   & &  &  &  &   \\ 
 & &   &  &  & &  &  \\  
     &&&&&&&\\ 
     &&&&&&&\\
\cline{1-8}\cline{1-8}

\multirow{14}{*}{\rotatebox[origin=c]{90}{\textsc{\textbf{Frequency Artifacts}}}} 
    & \multirow{5}{*}{\begin{tabular}[r]{@{}r@{}}\textbf{CF \#9. }\textit{ConvNet}\\\textit{Models}\end{tabular}} & \multirow{5}{*}{\begin{tabular}[c]{@{}l@{}}\textit{Frequency Learning} \\ \textit{Single-center Loss} \\ \textit{Phase Spectrum Learning} \\ \textit{Spatial \& Frequency Learning} \end{tabular}}  &  \multirow{5}{0.32\linewidth}{\includegraphics[trim={0pt 12pt 0pt 14pt},clip, width=1\linewidth,valign=b,page=26]{Figures/ConceptualFrameworkCategory.pdf}}
    & \multirow{5}{*}{\begin{tabular}[c]{@{}l@{}}\textsc{eccv} \\ \textsc{cvpr} \\ \textsc{cvpr} \\ \textsc{aaai} \end{tabular}} 
    & \multirow{5}{*}{\begin{tabular}[c]{@{}l@{}}\textquotesingle20 \\ \textquotesingle21 \\ \textquotesingle21 \\ \textquotesingle23 \end{tabular}} 
    & \multirow{5}{*}{\begin{tabular}[c]{@{}l@{}}F3-Net\cite{qian2020f3net} \\ FDFL\cite{li2021fdfl} \\ SPSL\cite{liu2021spsl} \\ LRL\cite{chen2021lrl} \end{tabular}} 
    & \multirow{5}{*}{\begin{tabular}[c]{@{}l@{}}98.10 \textsc{(auc)} \\ 97.13 \textsc{(auc)} \\ 95.32 \textsc{(auc)} \\ 99.46 \textsc{(auc)} \end{tabular}} \\
    &&&&&&&\\ 
     &&&&&&&\\
     &&&&&&&\\ 
     &&&&&&&\\
     &&&&&&&\\
\cline{2-8}
    & \multirow{4}{*}{\begin{tabular}[l]{@{}l@{}}\textbf{CF \#10. } \textit{ConvNet with}\\\textit{Sequence Model \&}\\\textit{Learning Strategies}\end{tabular}}
    &  \multirow{3}{*}{\begin{tabular}[c]{@{}l@{}}\textit{SpatioTemporal Frequency}\\\textit{Knowledge Distillation}\end{tabular}} 
    &  \multirow{3}{0.32\linewidth}{\includegraphics[trim={0pt 10pt 0pt 17pt},clip, width=1\linewidth,valign=b,page=29]{Figures/ConceptualFrameworkCategory.pdf}} 
     & \multirow{3}{*}{\begin{tabular}[c]{@{}l@{}}\textsc{eccv}\\\textsc{aaai}\end{tabular}} 
     & \multirow{3}{*}{\begin{tabular}[c]{@{}l@{}}\textquotesingle20\\\textquotesingle22\end{tabular}} 
     & \multirow{3}{*}{\begin{tabular}[c]{@{}l@{}}TRN\cite{masi2020trn}\\ADD$^\dagger$\cite{ble2022add}\end{tabular}} 
     & \multirow{3}{*}{\begin{tabular}[c]{@{}l@{}}\textsc{99.12 \textsc{(auc)}}\\95.46 \textsc{(acc)}\end{tabular}}\\
     &&&&&&&\\ 
     &&&&&&&\\
     &&&&&&&\\ 
     &&&&&&&\\
     \cline{2-8}
     & \multirow{4}{*}{\begin{tabular}[l]{@{}l@{}}\textbf{CF \#11. } \textit{Specialized }\\\textit{Network \& Learning}\\\textit{ Strategies}\end{tabular}}
          & \multirow{3}{*}{\begin{tabular}[c]{@{}l@{}}\textit{Intra-Sync with Frequency}\\\textit{Collaborative Learning} \end{tabular}}
     & \multirow{3}{0.32\linewidth}{\includegraphics[trim={0pt 10pt 0pt 16pt},clip, width=1\linewidth,valign=b,page=32]{Figures/ConceptualFrameworkCategory.pdf}} 
     & \multirow{3}{*}{\begin{tabular}[c]{@{}l@{}}\textsc{eccv}\\\textsc{cvpr} \end{tabular}}
     & \multirow{3}{*}{\begin{tabular}[c]{@{}l@{}}\textquotesingle22\\\textquotesingle23 \end{tabular}}
    & \multirow{3}{*}{\begin{tabular}[c]{@{}l@{}}CD-Net\cite{song2022cdnet}\\SFDG\cite{wang2023sfdg}\end{tabular}}  
     & \multirow{3}{*}{\begin{tabular}[c]{@{}l@{}}\textsc{98.50 \textsc{(auc)} }\\\textsc{95.98 \textsc{(auc)}}\end{tabular}}\\
      &&&&&&&\\
      &&&&&&&\\
      &&&&&&&\\
      &&&&&&&\\

\cline{1-8}\cline{1-8}

\multirow{9}{*}{\rotatebox[origin=c]{90}{\textsc{\textbf{Special Artifacts}}}}
    & \multirow{6}{*}{\begin{tabular}[r]{@{}r@{}}\textbf{CF \#12. } \textit{ConvNet}\\\textit{Models}\end{tabular}} 
    & \multirow{6}{*}{\begin{tabular}[c]{@{}l@{}}\textit{Region Tracking}\\\textit{Facial Features Modeling}\\\textit{2nd Order Anomaly}\\\textit{Audio-video Anomaly}\\\textit{Grad Pattern Learning}\end{tabular}}
     &  \multirow{6}{0.32\linewidth}{\includegraphics[trim={0pt 12pt 0pt 16pt},clip, width=1\linewidth,valign=b,page=35]{Figures/ConceptualFrameworkCategory.pdf}} 
     & \multirow{6}{*}{\begin{tabular}[c]{@{}l@{}}\textsc{cvpr}\\\textsc{cvpr}\\\textsc{cvpr}\\\textsc{cvpr}\\\textsc{cvpr}\end{tabular}}
     & \multirow{6}{*}{\begin{tabular}[c]{@{}l@{}}\textquotesingle21\\\textquotesingle21\\\textquotesingle22\\\textquotesingle23\\\textquotesingle23\end{tabular}}
    & \multirow{6}{*}{\begin{tabular}[c]{@{}l@{}}RFM\cite{wang2021rfm}\\FD2Net\cite{zhu2021fd2net}\\SOLA\cite{fei2022sola}\\AVAD\cite{feng2023avad}\\LGrad$^\dagger$\cite{tan2023lgrad}\end{tabular}} 
     & \multirow{6}{*}{\begin{tabular}{@{}c@{}}99.97 \textsc{(auc)} \\99.68 \textsc{(auc)}\\98.10 \textsc{(auc)}\\-\\\textsc{66.70 \textsc{(acc)}}\end{tabular}}\\
     &&&&&&&\\
     &&&&&&&\\
     &&&&&&&\\
     &&&&&&&\\
     &&&&&&&\\
\cline{2-8}
    & \multirow{4}{*}{\begin{tabular}[l]{@{}l@{}}\textbf{CF \#13. } \textit{Sequence }\\\textit{Model with Learning}\\ \textit{Strategies}\end{tabular}}
         & \multirow{3}{*}{\begin{tabular}[c]{@{}l@{}}\textit{Temporal Landmark Learning}\\\textit{Noise Pattern Learning}\end{tabular}}
     & \multirow{3}{0.32\linewidth}{\includegraphics[trim={0pt 10pt 0pt 17pt},clip, width=1\linewidth,valign=b,page=38]{Figures/ConceptualFrameworkCategory.pdf}}
     & \multirow{3}{*}{\begin{tabular}[c]{@{}l@{}}\textsc{cvpr}\\\textsc{aaai}\end{tabular}}
     & \multirow{3}{*}{\begin{tabular}[c]{@{}l@{}}\textquotesingle21\\\textquotesingle23 \end{tabular}}
     & \multirow{3}{*}{\begin{tabular}[c]{@{}l@{}}LRNet$^\dagger$\cite{sun2021lrnet}\\NoiseDF\cite{wang2023noisedf} \end{tabular}}
     & \multirow{3}{*}{\begin{tabular}[c]{@{}l@{}}\textsc{99.90 \textsc{(auc)}}\\\textsc{93.99 \textsc{(auc)}}\end{tabular}}\\
      &&&&&&&\\
      &&&&&&&\\
      &&&&&&&\\
      &&&&&&&\\
\cline{1-8}\cline{1-8}
\end{tabularx}
}

\end{table*}

\noindent
{\ballnumber{4} \textbf{Model \& Training. }}
The fourth step in our framework encompasses different choices of model architectures and training strategies commonly employed for deepfake detection. Our framework classifies the structure of the model into three broad categories of IFs: \pinkballnumbersmall{4A} \textit{Convolutional Models}, \pinkballnumbersmall{4B} \textit{Sequence Models}, and \pinkballnumbersmall{4C} \textit{Specialized Network}.

\textit{Convolutional Models} leverage common Convolutional Neural Networks (ConvNets) such as ResNet, VGG, or XceptionNet, which discern authentic images from manipulated ones by identifying subtle inconsistencies and anomalies in pixel patterns and textures.

\textit{Sequence Models} use Recurrent Neural Network (RNN) or Transformer-based model architectures, like BiLSTM, Vision Transformer, or Transformer Encoder, to analyze sequential inconsistencies. Spatiotemporal models track the continuity and flow of video frames to identify deepfakes. Alternatively, spatial detectors divide a single frame into multiple patches and input these as a sequence to the detector. 

\textit{Specialized Networks} models differ from the convolutional models category by integrating novel architectures such as U-Net~\cite{ronneberger2015u} or Capsule Network~\cite{sabour2017dynamic}, to capture more nuanced deepfake indicators.
Finally, Step 4 also includes \pinkballnumbersmall{4D} \textit{Learning Strategies} for training, such as meta-learning~\cite{chen2022ost}, Graph Information Interaction layers~\cite{wang2023sfdg}, Dual Cross-Modal Attention~\cite{luo2021frdm}, and Siamese learning~\cite{bonettini2021b4att}.

\noindent
{\ballnumber{5} \textbf{Model Validation.}}
Our fifth and final step of the framework addresses the critical task of validating pre-trained detectors. Based on our literature study, this validation process can be broadly categorized into two distinct approaches: the \coralballnumbersmall{5A} \textit{same dataset} and \coralballnumbersmall{5B} \textit{cross dataset} validation.

\textit{Same dataset} validation involves assessing the model's performance on the test set of the same dataset(s) as the training data (e.g., both training and test sets taken from FF++).

\textit{Cross dataset}, in contrast, involves testing the detector on a dataset different from the one from which the training dataset was taken (e.g., training data taken from FF++ and test data from CelebDF). This evaluation method is vital for assessing the model's generalizability and robustness.

\subsection{Detector Taxonomy}
\label{sub:taxonomy}

{In this section, we systematically categorize the \textbf{51} detectors identified through our selection process outlined in Section~\ref{sub:PaperSelectionCriteria}. We map each detector into a unified taxonomy based on our Conceptual Framework (\textbf{CF}), as presented in Table~\ref{tab:detectors-methodology}.}

Our analysis in Section~\ref{sub:DetectorsAnalysis} revealed that \textcolor[HTML]{7befe4}{\textbf{\textit{  Step \#2}}} and \textcolor[HTML]{ff83b4}{\textbf{\textit{Step \#4}}} of \textbf{CF} are particularly significant in determining the nature and type of the detector, as they dictate the model architecture choices and the targeted artifact, collectively termed as  `Focus of Methodology'. Consequently, we group the detectors by identifying their commonalities in these 2 steps of the conceptual framework.

We observe that utilizing IFs in Detector Methodology (i.e., \aquacapsule{} in \textcolor[HTML]{7befe4}{\textbf{\textit{Step \#2}}} of \textbf{CF}) naturally categorizes the 51 detectors into \textbf{4} high-level groups (see column 1 under `Focus of Methodology' in Table~\ref{tab:detectors-methodology}).
Subsequently, we identify subgroups within each of these 4 high-level groups using the influential factors in Model \& Training (i.e., \pinkcapsule{} in \textcolor[HTML]{ff83b4}{\textbf{\textit{Step \#4}}} in \textbf{CF}). This yields a total of \textbf{13} distinct \textbf{CF} sub-groups across these 4 high-level groups (see column 2 under `Focus of Methodology' in Table~\ref{tab:detectors-methodology}).

{We visually represent each of the 13 \textbf{CF} groups using color-coded nodes (i.e., capsule shapes) in Table~\ref{tab:detectors-methodology} under `Conceptual Framework Representation' column.
A fully colored node indicates that every paper in the framework belongs to the specified category. In contrast, a white node means that none of the papers in that framework fit the category. A half-colored node signifies a mixed scenario: some papers in the framework belong to the category, while others do not. Overall, these 13 \textbf{CF} representations depict various clusters of detector methodology and evaluation found in leading research publications since 2019. }

Additionally, our review identified key features of each detector's architecture, detailed in Table ~\ref{tab:detectors-methodology}'s third column, alongside the 13 \textbf{CF} groups, offering an overview of deepfake detection trends to be further explored in subsequent sections.

\subsubsection{Group 1: Spatial Artifact} There are \textbf{21} detectors in this group. There are three important observations to highlight about the whole group.

 \textit{Observation 1: Lacking capability to detect interframe inconsistencies.} All detectors in this group focus only on the spatial data. This also means that all approaches in this group are single frame (i.e. IF \orangeballnumbersmall{3D}) models. Therefore, none of these methods are equipped to find temporal deepfake artifacts that arise from the interframe inconsistencies in deepfake videos. 
 
 \textit{Observation 2:  Focus on AOI.} All detectors in this group use some method of face extraction on video frames to make the detection model focus only on the area of interest (AOI). Only 6 out of 51 detectors do not employ face extraction, highlighting the importance of IF \orangeballnumbersmall{3C}.
 
 \textit{Observation 3:  Generalizability claims against majority deepfakes.} All 21 detectors in this group claim the capability to detect both faceswap (IF \redballnumbersmall{1B}) and reenactment (IF \redballnumbersmall{1C}) deepfakes. For instance, 19 out of 21 detectors claim this on the same dataset evaluation (\coralballnumbersmall{5A}), and 17 out of 21 detectors claim this for cross dataset evaluation (\coralballnumbersmall{5B}). Therefore, claiming generalizability over two primary categories of deepfakes. 

The spatial artifact models are sub-grouped into 5 distinct \textbf{CF} representations based primarily on their differences in  \textcolor[HTML]{ff83b4}{\textbf{\textit{Step \#4}}} (Model \& Training).

\textbf{CF} \#1 contains purely ConvNets (IF \pinkballnumbersmall{4A}) that do not utilize sequence models or any additional specialized networks or learning strategies. Detectors characteristics in \textbf{CF} \#1 include such techniques that increase the performance of ConvNet models while focusing on the spatial data, for instance, the use of multiple color spaces~\cite{xu2023mcx}, consistency loss~\cite{ni2022core}, capsule network~\cite{nguyen2019capsule}, and depth-wise convolutions~\cite{Rossler2019ICCV}.

\textbf{CF} \#2 takes a different direction than \textbf{CF} \#1 by focusing on specialized network (IF \pinkballnumbersmall{4C}) architectures. For instance, these methods focus more on architecture choices like the use of siamese training~\cite{bonettini2021b4att}, multi-attention losses~\cite{zhao2021multi}, and intra-instance consistency loss~\cite{sun2022dcl}, demonstrating that these successful techniques from other domains are equally applicable in deepfake detection.

\textbf{CF}s \#3 and \#4, similar to \textbf{CF} \#1, rely on ConvNet models. However, \textbf{CF} \#3 incorporates additional special learning strategies (IF \pinkballnumbersmall{4D}), such as adversarial learning~\cite{cao2021mlac} and meta-learning~\cite{chen2022ost}, to improve the model's learning capabilities in an effort to enhance the deepfake detection performance. In contrast, \textbf{CF} \#4 integrates specialized networks (IF \pinkballnumbersmall{4D}) with ConvNets to benefit from techniques like collaborative learning~\cite{woo2023qad} which provide (video) quality agnostic detection performance on both raw and compressed deepfakes.

\textbf{CF} \#5 is unique as it relies on purely sequential models (IF \pinkballnumbersmall{4B}). Sequential models are mainly used for temporal or spatiotemporal data. Therefore, the spatial data needs to be transformed into a sequence, achieved by slicing each video frame into smaller chunks and feeding them sequentially to the model. This setting is unique and also boosts the detection performance as it helps in identifying faces versus other region inconsistencies~\cite{dong2022ict} and also enables the use of unsupervised methods~\cite{zhuang2022uiavit} for deepfake detection.

\subsubsection{Group 2: Spatiotemporal Artifact} 
There were \textbf{15} spatiotemporal artifact-based detectors. \textbf{Three} key group observations are worth mentioning. 

\textit{Observation 1: No temporal-only detectors.} Due to the visual nature of facial deepfakes, all models in this category in addition to temporal artifacts (\aquaballnumbersmall{2B}) also focus on spatial artifacts (\aquaballnumbersmall{2A}). In fact, there is no temporal-only detector among all 51 detectors in this study.

\textit{Observation 2:  No-single frame detectors.} Building on the previous observation. The focus on the temporal aspect of the data means that all detectors use multiple frames from the video at once for deepfake detection. This aspect helps in identifying the interframe inconsistencies.

\textit{Observation 3: Challenge in finding the balance.} As the focus of detectors is now divided among two spaces, spatial and temporal, it becomes a challenge to find the perfect balancing point where the best detection performance could be obtained from the features of these two spaces.

All the spatiotemporal models additionally utilize face extraction in Step 3. Among the spatiotemporal detectors, there are three distinct conceptual framework representations.

\textbf{CF} \#6, similar to \textbf{CF} \#1, focuses purely on ConvNets (\pinkballnumbersmall{4A}) however due to the presence of temporal data the techniques employed in these detectors differs significantly from \textbf{CF} \#1. For instance, the focus is more toward methods which take advantage of temporal aspects like time discrepancy learning~\cite{zhang2021td3dcnn}, anomalies in heartbeat rhythm~\cite{qi2020deeprhythm}, multi-instance learning~\cite{li2020simlt} and global-local frame learning~\cite{hu2021dia}.

\textbf{CF} \#7 is the most diverse among all CFs in its selection of \textcolor[HTML]{ff83b4}{\textbf{\textit{Step \#4}}} IFs, as it constitutes ConvNet detectors (\pinkballnumbersmall{4A}) with a mix of sequence model (\pinkballnumbersmall{4B}), specialized network (\pinkballnumbersmall{4C}) and learning strategies (\pinkballnumbersmall{4D}), in an effort to learn spatiotemporal inconsistencies or physiological behaviors like reading mouth movements. There are only limited studies that have explored this avenue, making it a promising area for future research.

\textbf{CF} \#8 contains exclusively sequence model-based detectors (\pinkballnumbersmall{4B}), which is the most straight forward choice for temporal data. However, due to the spatial nature of facial deepfakes as discussed earlier, techniques like spatio-temporal modules in convolutional LSTMs~\cite{tariq2021clrnet} and vision transformers~\cite{coccomini2022ccvit} become more relevant in this context. As they help the model learn spatiotemporal features of the data better than typical sequence models such as LSTMs and transformers.

\subsubsection{Group 3: Frequency Artifact} \label{sec:frequency} This group comprises \textbf{8} frequency-based detectors, highlighted by \textbf{2} key observations.

\textit{Observation 1: Few frequency-only detectors.} There are only \textbf{2} detectors that are purely targeting frequency artifact (\aquaballnumbersmall{2C}) whereas the remaining \textbf{6} target frequency with either spatial (\aquaballnumbersmall{2A}) or spatiotemporal (\aquaballnumbersmall{2A} \aquaballnumbersmall{2B}) artifacts. This shows that the frequency features are mostly considered as supplementary information just to enhance the detection performance and more emphasis is on the visual aspect through spatial or spatiotemporal data. 

\textit{Observation 2: No frequency-temporal detectors.} The combination of frequency artifacts (\aquaballnumbersmall{2C}) with temporal artifacts (\aquaballnumbersmall{2B}) is not explored in any of the detectors. This may be due to the visual emphasis on spatial components in facial deepfakes, making them a primary focus for feature learning. However, the efficacy of this approach remains unconfirmed without additional research.

There are 3 distinct CFs from  frequency artifact group.

\textbf{CF} \#9 consists of 2 detectors that focus exclusively on frequency artifacts, and the other two, which also consider spatial artifacts. However, all 4 detectors use ConvNets (\pinkballnumbersmall{4A}) to learn the features of these artifacts, employing techniques like frequency learning~\cite{qian2020f3net}, phase spectrum learning~\cite{liu2021spsl} and spatial-frequency learning~\cite{chen2021lrl}. Their performance on same dataset (\coralballnumbersmall{5A}) validation showcase that ConvNets are equally good for both type of artifacts.

\textbf{CF} \#10 and \#11 both focus on spatiotemporal artifacts in addition to frequency. However, \textbf{CF} \#10 detectors use ConvNet models (\pinkballnumbersmall{4A}) with techniques like Knowledge distillation~\cite{ble2022add} whereas, \textbf{CF} \#11 detectors opt for more specialized networks (\pinkballnumbersmall{4C}) to use technique collaborative learning~\cite{wang2023sfdg} and intra-sync with frequency~\cite{song2022cdnet}. Overall both \textbf{CF} \#10 and \#11 target same artifacts while choosing significantly different methodologies.

\subsubsection{Group 4: Special Artifact} This group consists of \textbf{7} detectors, marked by one significant observation.

\textit{Observation: Beyond spatial, temporal and frequency artifacts.} This group highlight a unique but very significant aspect, i.e., targeting a mix of spatial, temporal and frequency artifacts is important in developing an effective deepfake detector. However, at the same time, targeting higher level characteristics, such as face region tracking~\cite{wang2021rfm}, audio-video anomalies~\cite{feng2023avad}, gradient patterns~\cite{tan2023lgrad}, temporal landmarks~\cite{sun2021lrnet}, and noise patterns~\cite{wang2023noisedf}, provide more meaningful and explainable features for deepfake detectors.

A limited amount of work has been done in the special artifact (\aquaballnumbersmall{2D}) category. We categorized them into two CFs.

\textbf{CFs} \#12 and \#13 target higher-level characteristics using different models: \textbf{CF} \#12 employs ConvNet models (\pinkballnumbersmall{4A}) while \textbf{CF} \#13 uses sequence models (\pinkballnumbersmall{4B}), showcasing diverse detection approaches for various deepfake manipulations. For example, temporal landmarks~\cite{sun2021lrnet} are better captured by sequence models, whereas ConvNets excel in facial feature modeling~\cite{zhu2021fd2net}. Selecting the appropriate Model \& Training methodology (\textcolor[HTML]{ff83b4}{\textbf{\textit{Step \#4}}}) hinges on the detector’s focus (i.e., special artifacts), identifiable through our 5-step conceptual framework and taxonomy.

\section{Evaluation Settings}
\label{sec:EvaluationSettings}
The current detectors research landscape poses challenges in comparing model performance due to variations in datasets, metrics, and methodologies, obscuring the impact of model architecture and training methods. To address this and \textbf{\textit{RQ2}}, we employ a systematic evaluation approach to streamline performance comparison and identify IFs from our detector taxonomy. We rigorously evaluated deepfake detectors on various datasets to ensure fairness. From the detectors identified in \textbf{\textit{RQ1}}, we carefully selected \textbf{16}, based on strict criteria (Section~\ref{sec:DetectorforEvaluation}). These detectors formed the basis for subsequent experiments. We chose evaluation strategies and datasets for three settings: gray-box, white-box, and black-box (Section~\ref{sec:Evaluation Strategies}).

\subsection{Detectors for Evaluation}
\label{sec:DetectorforEvaluation}

To address \textit{\textbf{RQ2}}, we rigorously evaluated the performance of various deepfake detectors across various datasets, ensuring a meticulous and equitable comparison. To this end, we selected a subset of detectors from the  {51} identified in \textit{\textbf{RQ1}} (see Table \ref{tab:detectors-methodology}) by employing the following inclusion criteria:

\textbf{(i) Generalization Claims.} Only detectors with demonstrated generalizability on unseen datasets were selected, focusing on those explicitly designed for broad applicability across deepfake variants.

\textbf{(ii) Open Source with Model Weights.} Given the difficulty of replicating training environments, we included only open-source detectors with available pre-trained models, which resulted in 16 SoTA detectors that are indicated by $\dagger$ in Table \ref{tab:detectors-methodology}. These were mainly trained on the FF++ dataset, except for CLRNet and CCViT, which used DFDC, and their generalizability was tested on advanced deepfake datasets, as detailed in subsequent sections.

\textbf{Pre-training Sources.}
Most methods leverage pre-training on the FaceForensics++ datasets \cite{Rossler2019ICCV}, either partially or entirely. Notably, the CLRNet and CCViT models undergo pre-training using the DFDC dataset, specifically prepared for the Deepfake Detection Challenge; therefore, we omitted them from DFDC results in Fig.~\ref{fig:GrayBoxResults}. A distinct approach to pre-training is observed in LGrad, where the authors employ a novel dataset generated with ProGAN.

\textbf{Inference Process.}
During inference, detector configurations adhere strictly to the specifications in the respective paper. 
For frame-based prediction methodologies, we aggregate frame predictions to derive video probability or scores. Conversely, for multi-frame dectors, we selectively sample frames based on their designated temporal length for prediction. All inferences are run on a single NVIDIA GeForce RTX 3090 GPU.

\textbf{Evaluation Metrics.} In the main manuscript, we focus on the AUC and F1 score due to their resilience to class imbalance, and their frequent use in the literature. Additional metrics, including accuracy (ACC), recall, and precision, are detailed in Table \ref{tb:performance} in our Supplementary Material.

\subsection{Evaluation Strategies \& Datasets}
\label{sec:Evaluation Strategies}

\begin{table}
\centering
\caption{\textbf{Evaluation Strategy}. ``Detection Difficulty'' indicates the level of prior knowledge availability.}
\label{tb:dataset_cat}
\resizebox{1\linewidth}{!}{%
\begin{tabular}{@{}l|ccc|c@{}}
\toprule
\textbf{Datasets}&\parbox[c]{1.3cm}{\centering \textbf{Creation}\\\textbf{Control}} &\parbox[c]{1.9cm}{\centering \textbf{Source/Target}\\\textbf{Knowledge}} & \parbox[c]{1.5cm}{\centering \textbf{Method}\\\textbf{Knowledge}} & \parbox[c]{1.5cm}{\centering \textbf{Detection}\\\textbf{Difficulty}}\\ \hline\hline
White-box &\textcolor{green}{\cmark}           & \textcolor{green}{\cmark}            & \textcolor{green}{\cmark}  &   \textcolor{red}{$\bullet$} \textcolor{white}{$\bullet \bullet$}         \\
Gray-box &\textcolor{red}{\xmark}            & \textcolor{orange}{Partly}          & \textcolor{orange}{Partly}            &   \textcolor{red}{$\bullet  \bullet$} \textcolor{white}{$\bullet$}       \\
Black-box &\textcolor{red}{\xmark}            & \textcolor{red}{\xmark}           & \textcolor{red}{\xmark}              &   \textcolor{red}{$\bullet \bullet \bullet$}     \\
\bottomrule
\end{tabular}
}
\end{table}
Driven by \textbf{\textit{RQ2}} and \textbf{\textit{RQ3}}, we implement three evaluation strategies-black-box, gray-box, and white-box settings (Table \ref{tb:dataset_cat})-to reflect varying transparency and control over the deepfake generation process. This methodology addresses the gaps in prior surveys, which primarily focus on gray-box scenarios with limited exploration of black-box contexts.

    \textbf{Gray-box Generalizability Evaluation.} Our objective is to thoroughly evaluate datasets where we possess partial knowledge of, yet lack control over the source, destination, or generation method. By subjecting all detectors to gray-box settings, and employing two benchmark datasets: DFDC \cite{dolhansky2020deepfake} and CelebDF \cite{li2020celeb}, we aim to simulate scenarios where detectors have limited information about the deepfake generation process. This scenario represents a middle ground between black-box and white-box evaluations, where detectors operate with partial information, reflecting common real-world scenarios where some knowledge exists but complete control is lacking.

The DFDC dataset \cite{dolhansky2020deepfake} , released by Facebook, contains more than 100,000 faceswap videos of 3,426 actors, diverse in gender, age and ethnicity. Due to limited public information about its creation, it suits gray-box evaluation. The subsequent CelebDF dataset CelebDF \cite{li2020celeb} presents more sophisticated deepfakes with 590 original YouTube-sourced videos of celebrities with diverse demographics in terms of age, ethnicity, and gender, leading to 5,639 DeepFake videos. This enhances the variety of challenging samples for evaluation.
    
    \textbf {White-box Generalizability Evaluation.} Our study uniquely evaluates deepfake detectors in controlled environments, where we systematically control the video sources, targets, and the generation process. After initially identifying 20 leading tools as candidates, our rigorous selection criteria narrowed the choices down to 7, as detailed in  Table \ref{tab:criteria-4-apps-selection}). This evaluation setup aims to mimic scenarios where we have full information and control over the deepfake generation, providing insights into the detector's performance under different conditions. While \textbf{\textit{RQ2}} could potentially be addressed through validation in gray-box scenarios, the lack of transparency in gray-box settings in previous studies has hindered a thorough examination of \textbf{\textit{RQ3}}.  This white-box approach allows for an exhaustive assessment of Influential Factors (IFs) identified in \textbf{\textit{RQ1}}, which would be challenging in less transparent settings. 
    \begin{table}[h]
\centering
\caption{\textbf{Deepfake Generation Tools Included in the White-box Study.} Our selected deepfake generators are highlighted in green. In the table, the ``Being Serviced'' column indicates whether the program is still operational or outdated, with the last update year provided beside it.}
\label{tab:criteria-4-apps-selection}
\resizebox{\columnwidth}{!}{
\begin{tabular}{lccclc|c}
\hline\hline
\textbf{Program} &
\begin{tabular}[c]{@{}c@{}}\textbf{Open}\\\textbf{Source}\end{tabular}  & 
\textbf{Star No.} &
\textbf{Fork No.} &
\begin{tabular}[c]{@{}c@{}}\textbf{Being}\\\textbf{Seviced}\end{tabular}  & 
\begin{tabular}[c]{@{}c@{}}\textbf{Freedom of}\\ \textbf{Victim\&Driver}\end{tabular}  &
\begin{tabular}[c]{@{}c@{}}\textbf{Score}\\ \textbf{(max:6)}\end{tabular} \\ 
\hline \hline
FacePlay\cite{FacePlay}              & \xmark      & - & -  & \textcolor{green}\cmark   (2023) & \textcolor{green}\cmark~\xmark   & 2  \\ 
DeepFakesWeb \cite{DeepFakesWeb}     & \xmark      &  - & -  & \textcolor{green}\cmark    & \textcolor{green}\cmark~\textcolor{green}\cmark & 3 \\
\rowcolor[HTML]{ECFFDC}DeepFaceLab\cite{DeepFaceLab}     &\textcolor{green}\cmark& 42k  & 9.4k  & \textcolor{green}\cmark (2022)& \textcolor{green}\cmark~\textcolor{green}\cmark & \textbf{\textcolor{green}6}  \\ 
DeepFaceLive \cite{DeepFaceLive}   &\textcolor{green}\cmark & 17.1k  & 2.5k  & \textcolor{green}\cmark  (2023)  &  \xmark~\textcolor{green}\cmark & 5 \\ 
FaceApp \cite{FaceApp}               & \xmark      & - & -   & \textcolor{green}\cmark   (2023) & \textcolor{green}\cmark~\xmark   & 2 \\
Reface \cite{Reface}               & \xmark      & - & -   & \textcolor{green}\cmark   (2023) & \textcolor{green}\cmark~\xmark   & 2  \\
\rowcolor[HTML]{ECFFDC}Dfaker \cite{dfaker}              &\textcolor{green}\cmark & 461  & 151   & \textcolor{green}\cmark  (2020)  & \textcolor{green}\cmark~\textcolor{green}\cmark & \textbf{\textcolor{green}6} \\ 
\rowcolor[HTML]{ECFFDC}Faceswap \cite{faceswap}&\textcolor{green}\cmark& 46.7k & 12.6k  & \textcolor{green}\cmark (2023) & \textcolor{green}\cmark~\textcolor{green}\cmark & \textbf{\textcolor{green}6}  \\ 
\rowcolor[HTML]{ECFFDC}LightWeight \cite{faceswap}&\textcolor{green}\cmark& 46.7k & 12.6k  & \textcolor{green}\cmark (2023) & \textcolor{green}\cmark~\textcolor{green}\cmark & \textbf{\textcolor{green}6}  \\ 
deepfakes's faceswap \cite{deepfakes} &\textcolor{green}\cmark& 3k   & 1k   & \xmark   (2018) & \textcolor{green}\cmark~\textcolor{green}\cmark & 5 \\ 
Faceswap-GAN \cite{faceswap-GAN}   &\textcolor{green}\cmark& 3.3k  & 840  & \xmark  (2019) & \textcolor{green}\cmark~\textcolor{green}\cmark & 5 \\ 
\rowcolor[HTML]{ECFFDC}
FOM-Animation \cite{fom2019}       &\textcolor{green}\cmark& 13.7k  & 3.1k  & \textcolor{green}\cmark (2023) & \textcolor{green}\cmark~\textcolor{green}\cmark & \textbf{\textcolor{green}6} \\ 
\rowcolor[HTML]{ECFFDC}FOM-Faceswap \cite{fom2019}&\textcolor{green}\cmark& 13.7k & 3.1k  & \textcolor{green}\cmark  (2023) & \textcolor{green}\cmark~\textcolor{green}\cmark & \textbf{\textcolor{green}6} \\ 
\rowcolor[HTML]{ECFFDC}FSGAN \cite{nirkin2019fsgan}&\textcolor{green}\cmark& 702  & 143   & \textcolor{green}\cmark  (2023) & \textcolor{green}\cmark~\textcolor{green}\cmark & \textbf{\textcolor{green}6} \\ 
DeepFaker \cite{DeepFaker}          & \xmark        &  - & -   & \textcolor{green}\cmark (2023) & \textcolor{green}\cmark~\xmark  & 2 \\ 
Revive \cite{Revive}               & \xmark        &  - & -   & \textcolor{green}\cmark (2023) & \textcolor{green}\cmark~\xmark   & 2 \\ 
Fakeit \cite{Fakeit}              & \xmark        &  - & -   &  \xmark                 & \xmark~\xmark  & 0 \\ 
DeepFaker Bot \cite{DeepFakerBot}   & \xmark        &  - & -   & \xmark  & \textcolor{green}\cmark~\textcolor{green}\cmark & 2 \\ 
Revel.ai \cite{Revelai}            & \xmark        &  - & -  & \textcolor{green}\cmark  (2023) & \textcolor{green}\cmark~\textcolor{green}\cmark & 3  \\
SimSwap \cite{Chen_2020}           &\textcolor{green}\cmark& 2   & 703  & \textcolor{green}\cmark  (2023) & \textcolor{green}\cmark~\textcolor{green}\cmark  & 5 \\ 
licolico \cite{licolico}          & \xmark        & - & -   &   \xmark  (Closed)& \textcolor{green}\cmark~\xmark  & 1  \\ 
Deepfake Studio \cite{DeepfakeStudio}&\xmark       & - & - & \textcolor{green}\cmark  (2023)  & \textcolor{green}\cmark~\textcolor{green}\cmark  & 3\\ 
Deepcake.io \cite{Deepcakeio}    & \xmark        &  - & -   & \xmark  (Closed) & \xmark~\xmark  & 0\\ 
\hline\hline
\end{tabular}}
\end{table}

 Our comprehensive procedure for preparation and generation of white-box dataset is described as follows:

      \indent \quad \textbullet \, \textbf{{Selected Generators} }Table \ref{tab:criteria-4-apps-selection} summarizes all published deepfake creation tools covered in our survey. Columns two through six delineate our criteria for selecting deepfake creation tools to be included in our white-box dataset creation. Following a methodical evaluation process and the elimination of methods that did not meet our criteria, we curated a set of 7 distinct methods: DeepFaceLab~\cite{DeepFaceLab}, Faceswap~\cite{faceswap}, specifically the LightWeight variant within Faceswap, DeepFaker~\cite{DeepFaker}, FOM-Animation~\cite{fom2019}, and FSGAN~\cite{nirkin2019fsgan}. 
      
      \indent \quad \textbullet \, \textbf{Driver and Victim Video Selection}
We chose the real videos from the deepfake detection dataset (DFD) \cite{DFDGoogle} as the driver and victim videos for our deepfake video generation process for the following reasons: (i) All individuals featured in these videos are paid actors who have provided explicit consent for their videos to be utilized in deepfake generation for research purposes, and (ii) The dataset encompasses a wide variety of scenarios, enhancing its diversity in this context.

\indent \quad \textbullet \,   \textbf{Deepfake Generation Process}
By rigorously following a systematic process, we produced deepfake videos for each of the 7 selected generation methods. To generate these dataset videos, we conducted the following steps: (i) Selection of two random actors from a pool of 28 actors, (ii) Matching the scenarios portrayed in the original videos to both the source and target actors, emphasizing crucial elements such as facial expressions, body posture, and non-verbal cues to augment the video quality, and (iii) Provision of these videos to the selected deepfake generation methods. {Note: In most generation methods, deepfakes are produced iteratively, with visual quality progressively improving. To ensure consistency, a coauthor manually reviewed the visual fidelity, terminating the process when no further improvement was observed after multiple iterations.}. The real (source and target) videos utilized for deepfake generation constitute the real segment of the dataset, comprising 54 videos. Meanwhile, the deepfake segment of the dataset encompasses 28 videos for each of the 7 distinct methods, resulting in a total of 196 videos (28 × 7 = 196). In aggregate, our stabilized dataset comprises 250 videos and with an average duration of 35 sec/video, our dataset yields up to 167,000 fake frames, providing robust basis for evaluating detectors in a white-box setting.

    \textbf {Black-box Generalizability Evaluation.} This evaluation setting prioritizes dataset assessments without any knowledge of the deepfake generation methods or their origins, mimicking real-world scenarios.  
    We assembled a comprehensive dataset from links provided by \cite{{cho23_RWDF}} comprising 2,000 samples sourced from 4 online platforms: Reddit, YouTube, Bilibili, and TikTok,  and annotated with different  
intentions, demographics, and contexts. The lack of information regarding deepfake generation methods aligns with the challenges akin to real-world detection scenarios, emphasizing the need for detectors to perform effectively under such conditions. Adopting method in \cite{zheng2021ftcn}, we extracted and labeled the first clip of each video, resulting in 513 genuine and 1,383 manipulated clips, excluding 104 clips due to false positives from the face extractor in static artwork.

\section{Evaluation Results}
\label{sec:evaluation}

This section outlines our results motivated from \textbf{RQ2} and \textbf{RQ3}. We begin by summarizing our initial observations across all datasets in Section~\ref{sec:experiments}. Subsequently, we explore how the conceptual framework impacts detector generalizability in Section~\ref{sub:inpact_factors}. Our evaluation of the performance of the chosen detectors primarily focuses on their AUC and F1 scores.

 \begin{figure*}  
  \centering
  \includegraphics[width=0.70\textwidth]{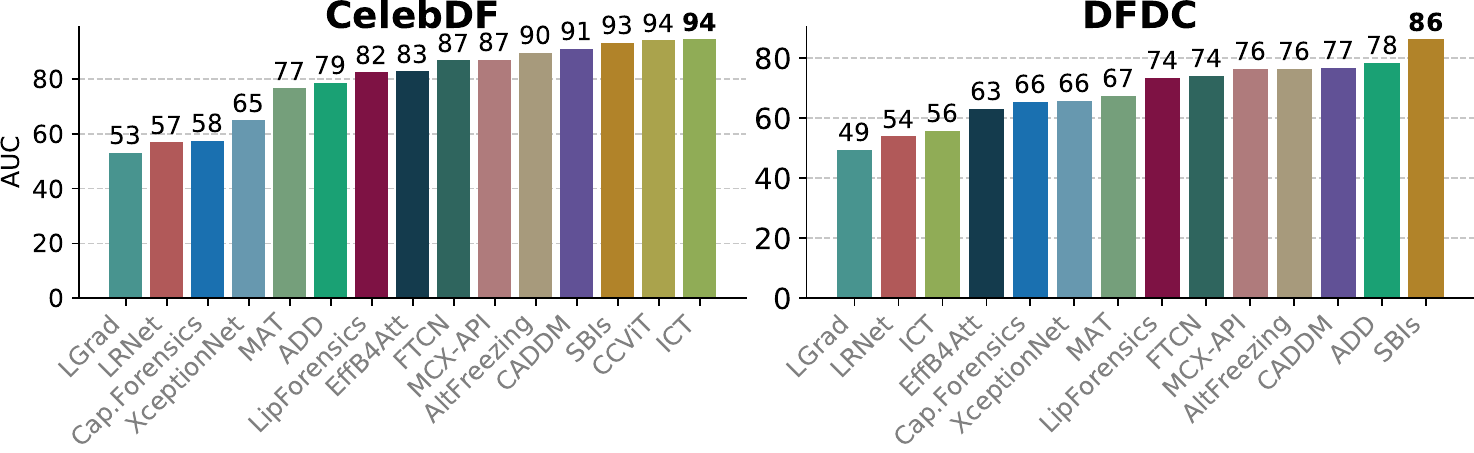}

  \caption{\includegraphics[scale=0.03]{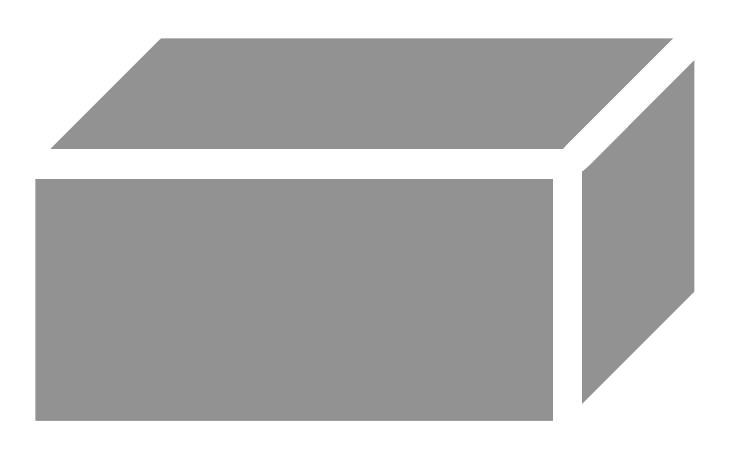} Gray-box results. Performance (AUC\%) of selected deepfake detectors on CelebDF and DFDC datasets. The overall performance of detectors on the DFDC dataset tends to be lower than CelebDF. To ensure a fair cross-evaluation comparison, we exclude CCViT since this model was trained directly on DFDC rather than the standard deepfake dataset, FF++.}
  \label{fig:GrayBoxResults}
 
\end{figure*}

\subsection{Detection Results}\label{sec:experiments} 
\subsubsection{Gray-box generalizability}\label{sec:gray-box}
This section presents generalizability results derived from the raw performance metrics on the CelebDF and DFDC datasets presented in Fig. \ref{fig:GrayBoxResults}.  Our key findings include:

\indent \quad \textbullet \, \textbf{Environmental factors hinder detectors' ability to even find obvious artifacts.} 
Environmental factors, such as lighting and video quality, play a crucial role in deepfake detection. While CelebDF showcases superior deepfake quality compared to DFDC, the pristine lighting and high-quality camera setups in CelebDF videos paradoxically make it easier for detectors to spot deepfake artifacts. Conversely, the poorly lit environments and lower-quality recordings in DFDC create challenges for detectors, making it harder to identify even significant deepfake artifacts. Despite both being second-generation benchmarks,  CelebDF~\cite{Celeb_DF_cvpr20} presents a greater challenge than DFDC, with detectors reporting lower average performance compared to DFDC. However, our study suggests a different perspective, where a subset of 10 detectors, characterized by increased diversity and recency, exhibited lower performance on DFDC than CelebDF. Notably, detectors like LGrad, LRNet, and Cap.Forensics showed consistently low performance on both datasets, ranging from the mid-50s to mid-60s, rendering them non-competitive. Although CelebDF was released after DFDC and offers more detailed information, 
the latter features more background noise and varied lighting conditions. Consequently, the average performance of the detectors in CelebDF (79.30\%) exceeds that of DFDC (68.72\%) by 10.58\%. Understanding and incorporating these auxiliary factors is crucial for enhancing detection performance. 

\indent \quad \textbullet \, \textbf{Identity-based methods are only successful when the target demographic is known.}
The ICT detector, trained on celebrity faces, performs well on CelebDF, demonstrating the effectiveness of identity-based methods when the target demographic matches the training data. However, ICT struggles on DFDC, which lacks celebrity faces. In contrast, identity exclusion strategies like CADDM excel across both gray-box datasets, proving their robustness and suitability for unknown demographics.

\indent \quad \textbullet \, \textbf{Spatiotemporal artifact models are consistent performers but mainly work with videos.} In both datasets, multiple frame-based detection methods (\orangeballnumbersmall{3E}) rank among the top performers, notably LipForensics, FTCN, and AltFreezing. Indeed, with the exception of SBIs and CADDM, single-frame-based methods (\orangeballnumbersmall{3D}) were found to exhibit strong performance on only one of the two datasets evaluated. In contrast, the aforementioned multiple frame-based methods demonstrate consistent performance across both gray-box datasets. However, they exhibit some limitations, such as slower evaluation times due to multiple frame processing and their limitation in single image detection due to missing temporal elements. Still, these spatiotemporal-based methods showcase promising potential and should be a future research direction in deepfake detection.

\indent \quad \textbullet \, \textbf{Having a well-rounded and sophisticated model architecture is still relevant and improves generalization.} The EfficientNet architecture excels in gray-box datasets, ranking as a top performer with SBIs achieving AUC scores of 93.2\% on CelebDF and 86.2\% on DFDC, while CADDM recorded 91.0\% and 76.8\%, respectively. Similarly, Transformer architectures like CCViT and ICT lead in CelebDF performance. These results underscore that superior architecture significantly boosts model robustness and generalizability, even with extensive training data, emphasizing the importance of architecture choice in detection methods.

\subsubsection{White-box generalizability}
\label{sub:white-box}
\begin{figure*}
    \centering
    \includegraphics[width=1\linewidth]{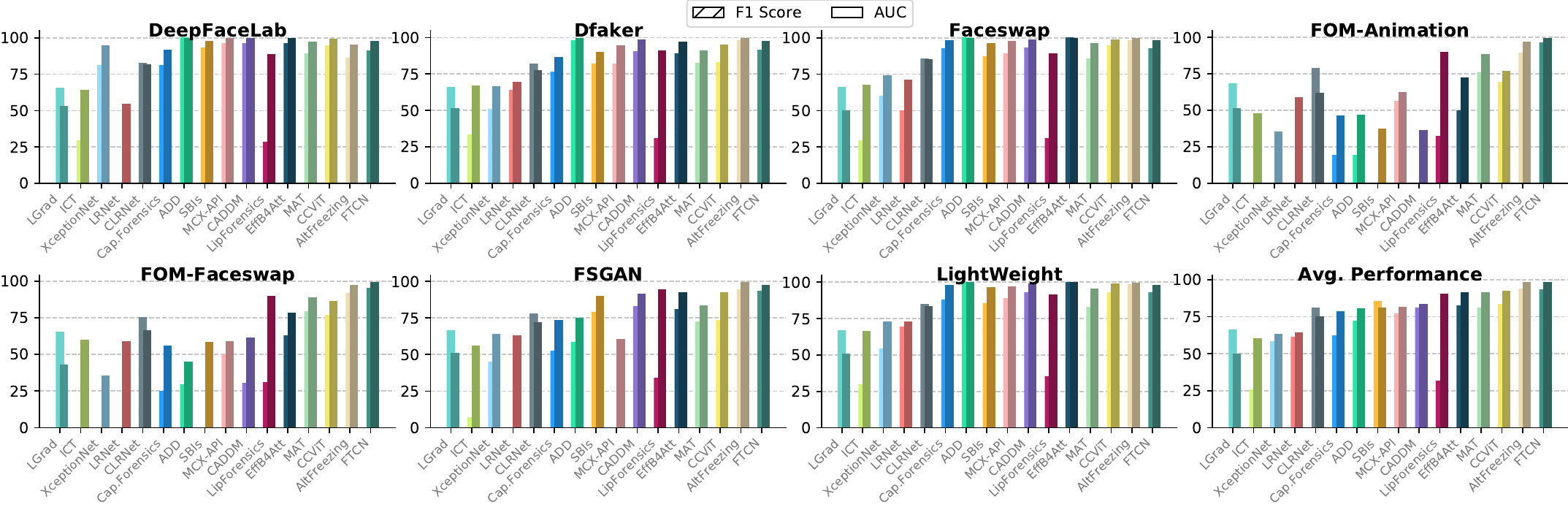}

    \caption{\includegraphics[scale=0.03]{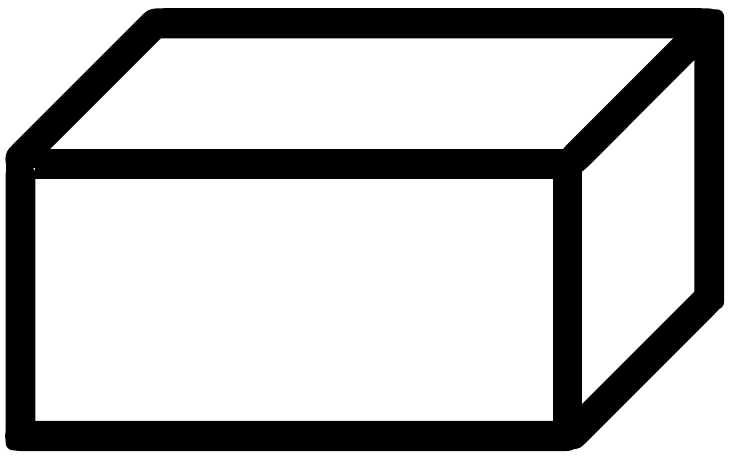} White-box results (Stabilized Set). F1 scores (dashed) and AUC scores (solid) of selected deepfake detectors.}
    \label{fig:whitebox}
\end{figure*}

Fig. \ref{fig:whitebox} shows the results of our white-box dataset experiments, leading to the following insights:

\indent \quad \textbullet \, \textbf{Spatial-temporal methods remain superior.} We observed that the two spatial-temporal models (\aquaballnumbersmall{2A}\aquaballnumbersmall{2B}), FTCN and AltFreezing, demonstrate the highest average performances, with scores of 98.4\% and 98.3\%, respectively. Remarkably, both methods achieved F1 and AUC scores above 80\% and 90\%, respectively, across all datasets. FTCN emphasizes learning temporal coherence, and is specifically designed to capture long-term coherence in videos. Conversely, AltFreezing is engineered as a spatiotemporal model, focusing on enhancing the forgery detection model's generalization capabilities. Furthermore, their performance does not significantly diminish when faced with geometric manipulation methods such as FOM (93\%) or prevalent deepfake techniques like DeepFaceLab (91\%).

\indent \quad \textbullet \, \textbf{Attention mechanisms contribute to enhanced detection performance.} Beyond multi-frame-based methods, attention-based approaches (\pinkballnumbersmall{4C}) demonstrate robust performance across datasets as the second-best category, surpassing an average AUC score of 91\%. CCViT leverages an attention mechanism derived from Transformer architecture (channel-wise attention), whereas MAT introduces a method to capture multiple face-attentional regions (spatial attention). Both achieved average AUC scores of 93\% and 92\%, respectively.  Similar to AltFreezing's experience, both CCViT and MAT excel in certain datasets but decrease on unseen videos from FOM methods (86\% and 89\%, respectively) and FSGAN (93\% and 83\%, respectively).

\indent \quad \textbullet \,\textbf{ Overfitting on very specific features and using simple loss functions leads to underperformance.}
 LGrad (47.78\%),  and ICT (61\%) are among the lowest average performers in our study. This underperformance is linked to each model's specialized focus. LGrad targets fully synthesized fake images (\redballnumbersmall{1A}) from GANs or Diffusion models, differing from the faceswap or reenactment scenarios common in our white-box tests. ICT, specifically designed for celebrity datasets like MS-Celeb-1M, relies on memorizing identities during training and utilizes ArcFace loss \cite{arcface2019} for identity comparison in face recognition tasks. This approach limits ICT's effectiveness on DFDC and white-box datasets, where the specific identities it has learned are absent. Such specialization, while beneficial in certain cases, can cause models to overfit to training data features, hindering their ability to generalize and detect a wide array of deepfakes.  Conversely, despite their specialized network designs, XceptionNet's sole reliance on basic optimization losses limits their generalization efficiency.


\begin{figure}
\centering
\includegraphics[width=0.95\linewidth]{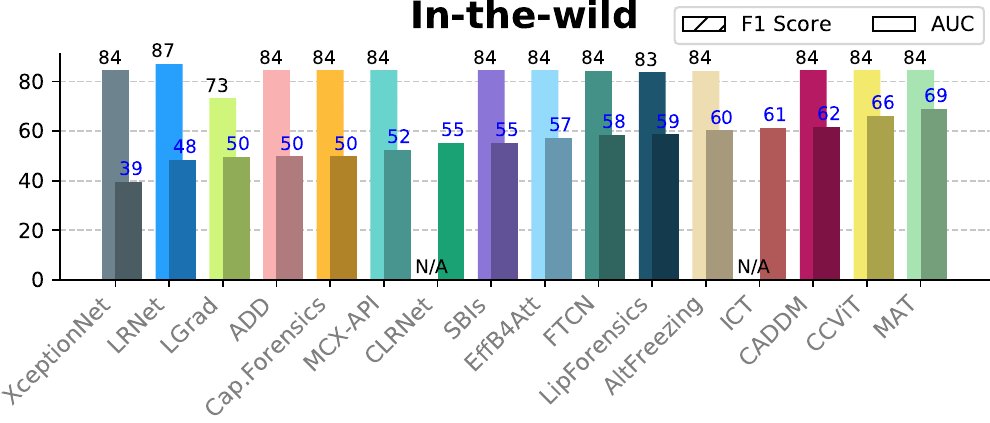}
\caption{\includegraphics[scale=0.03]{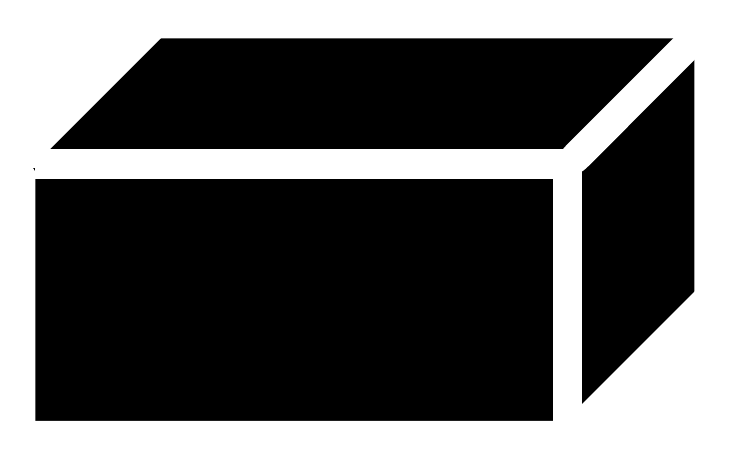} Black-box results (in-the-wild).}
    \label{fig:BlackBoxResults}
\end{figure}

\subsubsection{Black-box generalizability}

Our insights from the black-box experimental results in Fig. \ref{fig:BlackBoxResults} include:

\indent \quad \textbullet \, \textbf{Challenges and Discrepancies between F1 and AUC Scores in in-the-wild Deepfakes.} Despite F1 scores exceeding 84\%, no detector achieved an AUC over 70\%, highlighting the difficulty in detecting in-the-wild deepfakes. Some methods failed to identify any fakes among 1,383 clips at their optimal threshold, resulting in undefined (NaN) F1 scores. Conversely, a few other detectors fail to predict any of the pristine frames at their optimal threshold, leading to an absurdly high recall rate (see Table \ref{tb:performance} in Supplementary Material) yet impractical. Overall, we can observe such a big difference between the performance of AUC and F1 score in Fig. \ref{fig:BlackBoxResults}, highlighting how misleading it can be to use just a single metric.

\indent \quad \textbullet \, \textbf{Superiority of Attention-based and Multiple Frame-based Approaches.} The leading methods continue to be attention-based (MAT and CCViT) and multi-frame-based (AltFreezing, LipForensics, and FTCN) approaches, similar to findings in white-box evaluations. Given the prevalence of user-friendly software offering additional smoothing functions~\cite{cho23_RWDF}, detecting authenticity based solely on single-frame-based identification factors becomes challenging.
Nevertheless, despite its high AUC score, ICT struggles to effectively discern fake from genuine videos, as evidenced by a notable disparity between its AUC and F1 scores. 

\indent \quad \textbullet \, \textbf{Significance of network architecture.} Similar to the gray-box, three of the top detectors—MAT, CCViT, and CADDM— are based on spatial artifact detection. Despite methodological differences, they share common features: CADDM and MAT use EfficientNet.  Transformer-based architectures in CCViT also lead the field. This underscores the pivotal role of specialized network architectures, like EfficientNet and Transformers, over commonly reported networks, such as ResNet and XceptionNet, in detecting unknown deepfakes in the wild. Specifically, Transformers effectively model long-range dependencies in data, making them well-suited for identifying subtle patterns in deepfake content. Similarly, EfficientNets, utilizing network architecture search (NAS) and squeeze-and-excitation (SE) blocks, enhance manipulated facial regions while suppressing irrelevant ones.

\indent \quad \textbullet \, \textbf{Challenges for detectors without artifact mining or with overly specialized artifacts.} 
  Our analysis identifies XceptionNet  as less effective, with AUC scores of 39\% due to it reliance on basic optimization loss and lack of artifact mining, despite their specialized designs. On the other hand, methods targeting very specific artifacts, such as LRNet and LGrad, also underperform, indicating a deficiency in processing subtle spatial details and identifying nuanced anomalies like irregular blending or temporal inconsistencies in deepfakes.

\subsection{Impact of Influential Factors}
\label{sub:inpact_factors}

Our extensive evaluation using three distinctive settings demonstrates the impact of the identified IFs within our \textbf{CF}. To elucidate this impact, we provide specific illustrations without loss of generality:

(1) \textbf{The imperative of universality in deepfake detection}. Our classification of deepfake types into three distinct IFs under \textcolor[HTML]{ff4155}{\textbf{\textit{CF Step \#1}}} (Deepfake Type \redballnumbersmall{1A-C}) states that a fully developed deepfake detection method must be capable of effectively identifying new deepfakes belonging to these three groups. This underscores the significance of utilizing diverse data sources (Deepfake Type), and evaluation datasets as outlined in \textcolor[HTML]{ef6561}{\textbf{\textit{CF Step \#5}}} (Model Validation). A case in point is ICT, which is promising with CelebDF (\coralballnumbersmall{5A}) but exhibits significant underperformance on the DFDC dataset (\coralballnumbersmall{5B}), an entirely novel dataset for it, underscoring the hurdles in achieving broad generalization. Similarly, LGrad, which targets gradient artifacts in images generated by GANs and Diffusion models (\coralballnumbersmall{5A}), shows decreased efficacy in datasets featuring videos produced through Faceswap and reenactment techniques (\coralballnumbersmall{5B}). Therefore, a method's dependency on specific features from familiar training datasets may not suffice for the accurate detection of deepfakes in unfamiliar datasets. 

(2) \textbf{The significance of the detection methodology towards generalizability}. Our observations underscore that among the four IFs outlined in \textcolor[HTML]{7befe4}{\textbf{\textit{CF  Step \#2}}} (Detection Methodology), combining spatial (\aquaballnumbersmall{2A}) and temporal (\aquaballnumbersmall{{2B}}) elements enhances generalizability across contexts. Examples include AltFreezing and FTCN, which are effective because many deepfake generation techniques focus on manipulating individual frames, overlooking temporal coherence. Table \ref{app_tab:DetectorFurtherDetails} in Supp. \ref{sec:appendix-white-box} shows additional performance metrics for our white-box experiments, including recall rates. Analysis reveals that five models lead in recall performance, achieving an average of over 80\% recall across the white-box dataset while maintaining high AUC and F1: FTCN, AltFreezing, CLRNet, MAT and CCViT. Four of these five models target spatiotemporal artifacts (\aquaballnumbersmall{2A}\aquaballnumbersmall{2B}), except for MAT, which targets only spatial artifacts. In security-critical contexts like facial liveness verification, misclassifying a single deepfake as genuine poses significant risks. Thus, scrutinizing the recall metric, which indicates the percentage of deepfakes accurately identified by the model, is crucial.

On the other hand, factors lowering generalizability include heavy reliance on niche artifacts (\aquaballnumbersmall{{2D}}) by models like LRNet and LGrad, or ignoring artifact mining (using \aquaballnumbersmall{{2A}} alone) by XceptionNet, compromising model performance in unknown scenarios. Consequently, \textcolor[HTML]{fbbd5e}{\textbf{\textit{CF Step \#3}}} (Data Processing) is crucial for each approach's effectiveness. Using multiple frames (\orangeballnumbersmall{3E}) in spatial-temporal strategies, as in FTCN and AltFreezing, aids in learning generalizable deepfake indicators. Conversely, prioritizing image processing (\orangeballnumbersmall{3B}) to emphasize distinct artifacts, like landmarks in LRNet and gradients  in LGrad, may reduce effectiveness against novel deepfakes lacking these specific indicators.

(3) \textbf{Critical role of model architecture and learning approach.} In the challenging context of black-box scenarios, among the four IFs of  \textcolor[HTML]{ff83b4}{\textbf{\textit{CF Step \#4}}} (Model and Training), EfficientNet (such as MAT and CADDM) (\pinkballnumbersmall{4C}) and Transformer-based architectures (such as CCViT) (using a mix of \pinkballnumbersmall{4A-C}) emerge as the most effective, outperforming numerous alternatives. Additionally, attention-based learning strategies (\pinkballnumbersmall{4D}) prove exceptionally promising for both black and white-box environments, particularly the MAT method. Sequence models with spatial or temporal artifacts, as previously mentioned, show promise in most scenarios. Consequently, the considerations outlined in \textcolor[HTML]{ff83b4}{\textbf{\textit{CF Step \#4}}}—spanning all four categories—are essential for the development of a practical and more generalizable deepfake detector.
Common concerns regarding the IFs are discussed in Supp.~\ref{app_sec:moreIFs}.
\section{Discussion}
\label{sec:limit}

\subsection{Challenges in Reproducing SOTA} Examining over 50 deepfake detectors published in top venues from 2019 to 2023 reveals a concerning pattern. Only 15 (30\%) of these models have publicly released their pre-trained models. This lack of transparency, evident in the remaining 70\%, hampers reproducibility and limits understanding of their actual limitations, thereby obstructing effective comparative analysis. This accessibility issue slows down the evaluation of different methodologies, potentially hindering progress in deepfake detection. Promoting the release of pre-trained models is vital for enhancing comparative studies, accelerating advancements, and ensuring the robustness of these methodologies in real-world applications.

\subsection{Real-World Deepfake Detection is still an Open Issue} 
\begin{figure}[t]
    \centering
\frame{\includegraphics[width=0.99\linewidth]{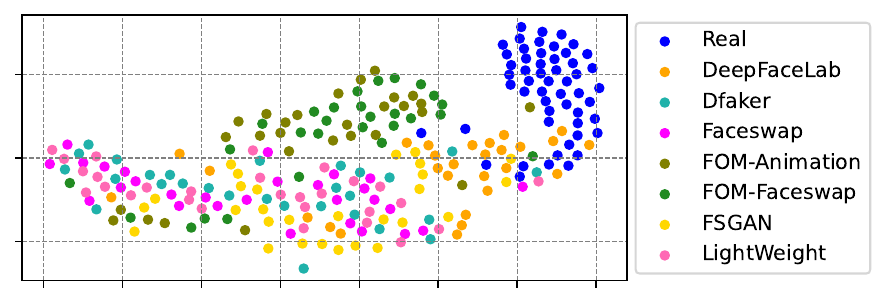}}
\caption{t-SNE of white-box datasets with AltFreezing method. Each dot corresponds to one video representation.}
    \label{fig:tsne_altfreezing}
\end{figure}

Our results reveal that no single detector consistently excels across all categories within our proposed three-tiered evaluation framework (black, gray, and white boxes). 
While many detectors claim to be generalizable based on gray-box evaluations, they are proficient mainly in specific scenarios. 
Specifically, detectors tailored for certain deepfake types, like faceswap (\redballnumbersmall{1B}) or reenactment (\redballnumbersmall{1C}), often falter when identifying other synthetic variants (\redballnumbersmall{1A}). 
Moreover, the difficulty of cross-dataset evaluation poses a significant challenge, potentially invalidating the generalizability claims of these detectors in the broader context of deepfake detection. 
To visually illuminate these distinctions, we employ dimensionality reduction via t-distributed stochastic neighbor embedding (t-SNE) to illustrate the divergent characteristics of samples from seven datasets, as perceived by the model (See Fig. \ref{fig:tsne_altfreezing}). We employ the AltFreezing model \cite{wang2023altfreezing} to process images from seven datasets and extract their intermediate representations. As shown, while the real dataset is distinctly separated from other deepfake types, some simpler fake types, such as FOM, are well differentiated, whereas others, like Lightweight and FSGAN, tend to overlap with the real video samples. Our findings demonstrate the necessity for a more comprehensive evaluation of generalizability, advocating for a thorough examination through our proposed evaluation strategy.

\subsection{Synthesis Deepfake Type is Overlooked}  Among the 51 detectors we examined, over 96\% primarily target reenactment or faceswap detection. Notably, the emergence of diffusion models activates synthesis-based deepfakes (\redballnumbersmall{1A}), yet research on effective detection mechanisms for them is limited. 
Although some initial efforts have been made to identify fully synthetic images generated by diffusion, this avenue is still in its infancy and requires substantial exploration~\cite{ricker2022towards}. The prospect of developing meta-detectors capable of distinguishing between faceswap, reenactment, and synthesis could serve as an initial step in the detection pipeline. 

\subsection{Influential Factors and Case Study on EfficientNet} 
While we meticulously identify the IFs that researchers consider in detector development and outline the impact of many as use cases, quantifying the individual influence of each factor on bottom-line efficacy proves challenging. 
This difficulty arises from the inherently predictive nature of AI-based models and the complexities in retraining detectors with diverse IF combinations, owing to insufficient construction details. 
Addressing this challenge remains an ongoing avenue for future exploration. Nevertheless, the comprehensive identification of these factors is invaluable, offering the potential to enhance the qualification of methods by leveraging a thorough understanding of these critical elements. This can benefit especially in uncovering their limitations. Therefore, we performed a case study (see Table \ref{tb:casestudy}) where we considered just one model under different IF settings that were feasible for the entire pipeline of that detector. We observed how the inclusion of different IFs impacted the performance across our evaluation strategies (i.e., gray box, white box, black box).

\begin{table}
\centering
\caption{Case Study of applying different Influential Factors (IFs) on EfficientNet model using our five-step conceptual framework. Here, 3A.1, 3A.2, and 3A.3 refer to different data augmentation of Auto Augmentation, Fake Augmentation, and Frequency Transform, respectively. Similarly, 3C.1, 3C.2, 3C.3, and 3C.4 refer to different face extraction methods such as MTCNN, Blazeface, Retinaface, and dlib. Also, 4C.1 and 4C.2 refer to the different settings of EfficientNet, such as 380 and 224, respectively, which were selected based on the compatibility with the other IFs. Lastly, 4D.1,  4D.2, and 4D.3 refer to different learning strategies such as ID-unaware, Siamese and Multi-attention.}
\label{tb:casestudy}
\resizebox{\linewidth}{!}{%
\begin{tabular}{cc|cc|ccc|cc|c|ccc|c} 
\toprule
\multicolumn{10}{c|}{\textbf{\textsc{Influential Factors}}} & \multicolumn{4}{c}{\textbf{\textsc{AUC Scores}}} \\
\hline
\multicolumn{2}{c|}{\textbf{Step \#1}} & \multicolumn{2}{c|}{\textbf{Step \#2}} & \multicolumn{3}{c|}{\textbf{Step \#3}} & \multicolumn{2}{c|}{\textbf{Step \#4}} & \textbf{Step \#5} & \textbf{White} & \textbf{Black} & \textbf{Gray} & \textbf{Avg.} \\ 
\hline

\redballnumbersmall{1B} & \redballnumbersmall{1C} & \aquaballnumbersmall{2A} &  & \orangeballnumbersmall{3A.1} & \orangeballnumbersmall{3C.1} & \orangeballnumbersmall{3D} & \pinkballnumbersmall{4C.1} &  & \coralballnumbersmall{5B} & 61.8 & 63.4 & 80.1 & 68.4 \\

\redballnumbersmall{1B} & \redballnumbersmall{1C} & \aquaballnumbersmall{2A} &  & \orangeballnumbersmall{3A.2} & \orangeballnumbersmall{3C.4} & \orangeballnumbersmall{3D} & \pinkballnumbersmall{4C.1} &  & \coralballnumbersmall{5B} & 81.1 & 55.3 & 89.7 & 75.4 \\

\redballnumbersmall{1B} & \redballnumbersmall{1C} & \aquaballnumbersmall{2A} & \aquaballnumbersmall{2C} & \orangeballnumbersmall{3A.3} & \orangeballnumbersmall{3C.1} & \orangeballnumbersmall{3D} & \pinkballnumbersmall{4C.1} &  & \coralballnumbersmall{5B} & 66.8 & 48.2 & 73.8 & 62.9 \\

\redballnumbersmall{1B} & \redballnumbersmall{1C} & \aquaballnumbersmall{2A} &  &  & \orangeballnumbersmall{3C.1} & \orangeballnumbersmall{3D} &   \pinkballnumbersmall{4C.1} &  & \coralballnumbersmall{5B} & 85.8 & 64.2 & 81.6 & 77.2 \\

\redballnumbersmall{1B} & \redballnumbersmall{1C} & \aquaballnumbersmall{2A} &  &  &  \orangeballnumbersmall{3C.1} & \orangeballnumbersmall{3D} &\pinkballnumbersmall{4C.2} &  & \coralballnumbersmall{5B} & 87.3 & 60.8 & 84.5 & 77.5 \\

\redballnumbersmall{1B} & \redballnumbersmall{1C} & \aquaballnumbersmall{2A} &  &  & \orangeballnumbersmall{3C.1} & \orangeballnumbersmall{3D} & \pinkballnumbersmall{4C.2} & \pinkballnumbersmall{4D.1} & \coralballnumbersmall{5B} & 83.7 & 61.5 & 83.9 & 76.4 \\
\redballnumbersmall{1B} & \redballnumbersmall{1C} & \aquaballnumbersmall{2A} &  &  &  \orangeballnumbersmall{3C.2} & \orangeballnumbersmall{3D} &\pinkballnumbersmall{4C.2} & \pinkballnumbersmall{4D.2} & \coralballnumbersmall{5B} & 91.5 & 57.4 & 81.5 & 76.8 \\
\redballnumbersmall{1B} & \redballnumbersmall{1C} & \aquaballnumbersmall{2A} &  &  &  \orangeballnumbersmall{3C.3} & \orangeballnumbersmall{3D} &\pinkballnumbersmall{4C.2} & \pinkballnumbersmall{4D.3} & \coralballnumbersmall{5B} & 91.6 & 68.9 & 81.8 & 80.8 \\

\bottomrule
\end{tabular}
}
\end{table}

\subsection{Ethical Considerations} 
\label{sub:ethical}
We emphasize ethical practices in creating and using deepfake datasets, including tools for creation and detection from the research community without any offensive content. Our approach has been approved by the ethics review boards of our organizations, reflecting our commitment to maintaining high ethical standards in our research.

\section{Future Directions}
\label{sec:future_direction}
We propose \textit{seven} strategic directions to combat the proliferation of deepfakes effectively.

\noindent
{\ballnumber{a} \textbf{\textbf{Conceptual Framework Utilization.} }} Developers seeking to create new deepfake detectors can utilize our Conceptual Framework (CF) as a strategic tool to validate their hypotheses and pinpoint the Influential Factors (IFs) required for achieving peak performance. This structured approach provides a roadmap for identifying and incorporating critical elements into detector design. For instance, those considering the adoption of transformer technology for deepfake identification can compare their initial hypothesis with the transformer-based methods (i.e.,  CCViT and ICT) identified by our CF to gain insights into their strengths and limitations in deepfake detection context, potentially streamlining the development process and saving significant effort.

\noindent
{\ballnumber{b} \textbf{\textbf{Adoption of Open Detectors and Three-Level Evaluations.} }} In this paper, we have laid out our recommended model evaluation framework that should be adopted in future deepfake detector studies. This includes subjecting them to thorough evaluation using gray-box, white-box, and black-box assessments and reporting results using extensive metrics, including AUC, precision, recall, and F1 score, in addition to accuracy. We additionally advocate for researchers to release their developed detector models. This approach ensures the validation of generalizability, promoting transparency and reliability in deepfake detection.

\noindent
{\ballnumber{c} \textbf{\textbf{Deeper Analysis of IFs.} }} The combination of our detector taxonomy and evaluation framework opens avenues for uncovering deeper insights into IFs. This can be achieved through meticulous, fine-grained ablation studies, employing consistent training hyperparameters and architectural settings, possibly enhanced by using white-box datasets for training.

\noindent
{\ballnumber{d} \textbf{\textbf{Multimodal and Specialized Model.} }} Future research should move beyond single-source data dependence, exploring multimodal models that integrate cues from audio, language, visual elements, and metadata (including identity). This comprehensive approach harnesses the synergistic effect of combining multiple data types, capitalizing on the strengths of various architectures and learning methodologies (\pinkballnumbersmall{4A-D}). Hence, it can significantly improve detection accuracy and robustness. On the other hand,  as discussed in Sec. \ref{sec:evaluation}, we found that network architecture plays a pivotal role in detection efficacy, despite initially being designed for other applications. Therefore, employing strategies like Neural Architecture Search (NAS) with reinforcement learning to discover optimal architectures (\pinkballnumbersmall{4C}) specifically tailored for deepfake detection represents a promising research avenue.

\noindent
{\ballnumber{e} \textbf{\textbf{Development of more resilient SoTA.} }}
Actively identifying deepfakes in various settings is crucial for developing robust detectors. We can establish a more rigorous testing environment by refining evaluation datasets to include malicious deepfakes that might be missed by searching online media platforms. Expanding the dataset to include content in different languages also widens detection capabilities. Updating training datasets with the latest techniques is essential to keep pace with emerging, more complex deepfakes. Also, incorporating continual and lifelong learning methods into evaluation and training ensures that detectors remain versatile and effective against dynamic threats posed by deepfakes. 

\noindent
{\ballnumber{f} \textbf{\textbf{Holistic Approach.} }} Effectively addressing the deepfake challenge requires a multi-faceted approach. This includes the integration of advanced detection technologies, data provenance tracking methods, comprehensive public education to raise awareness, and robust government policies to regulate usage. By synthesizing these diverse strategies, we can establish a resilient and comprehensive defense against the manipulation and misuse of deepfake technology.

\noindent
{\ballnumber{g} \textbf{\textbf{Proactive Rather Than Reactive.} }} 
A key research direction is to transition from solely reactive deepfake detection to pioneering proactive strategies, such as developing fingerprinting techniques for deepfake media. This allows for the tracing of origins and tracking of deepfake sources, lessening the dependence on broad-spectrum deepfake detection (\redballnumbersmall{1A-C} and \coralballnumbersmall{5A-B}). Implementing proactive defense strategies reduces the need for creating exhaustive model architectures and extracting specific or generalized artifacts (\orangeballnumbersmall{3A-B}) for newly emerging deepfakes, facilitating their early removal and curtailing the spread of misinformation.
\section{Social Impacts and Concluding Thoughts}
\label{sec:conclusion}
We believe deepfakes are becoming a more serious threat to our society, as they continue to grow in scale, complexity, and sophistication. There is an immediate need to examine various deepfake detection tools and understand their limitations through thorough analysis and evaluation to protect our society. Our work fills this gap with a detailed framework and assessment of current research, identifying spatiotemporal models such as FTCN as leaders but noting a significant disparity between claimed and actual detector performance. Future research should aim to develop more generalized detection methods, evaluate gray, black, and white-box settings, and explore proactive defenses. We also intend for our framework and evaluation methodology to adapt to new deepfake challenges.

\section*{Acknowledgement}

    This work was partly supported by Institute for Information \& communication Technology Planning \& evaluation (IITP) grants funded by the Korean government MSIT: (RS-2022-II221199, RS-2024-00337703, RS-2022-II220688, RS-2019-II190421, RS-2023-00230\\337, RS-2024-00356293, RS-2022-II221045, RS-2021-II212068, and RS-2024-00437849).

\bibliographystyle{plain}
\bibliography{egbib}

\newpage
\setcounter{page}{1}

\appendices
\hypersetup{linkcolor=black}
\startcontents[appendices] 
\printcontents[appendices]{}{0}{} 
\hypersetup{linkcolor=red} 
\renewcommand{\thesection}{\Alph{section}}

\section{Further Details on Detectors}
\label{app_sec:detector_further_details}
\begin{table*}[t]
\centering
\caption{\textbf{Further Details on detectors.} F2F, NT, and FOM stands for Facial reenactment methods Face2Face and Neural Textures. FS, DF, FSGAN, and FaSh stand for face swap methods FaceSwap, DeepFake, FaceswapGAN, and FaceShifter. }
\label{app_tab:DetectorFurtherDetails}
\resizebox{1\linewidth}{!}{%
\begin{tabular}{llllll} 
\hline\hline
 \textbf{\textcolor{black}{Paper Name}} & \textbf{\textcolor{black}{Deepfakes for Training}} & \textbf{\textcolor{black}{Artifacts}} & \textbf{\textcolor{black}{Data Pre-processing}} & \textbf{\textcolor{black}{Model Training}} & \textbf{\textcolor{black}{Model Validation}} \\ 
\hline\hline
\textbf{Cap.Forensics} & F2F, DF & Spatial & Single Frame & VGG & F2F, DF \\
 \textbf{XceptionNet} & NT, F2F, FS, DF & Spatial & dlib,Single Frame & XceptionNet & NT, F2F, FS, DF \\
\textbf{Face X-ray} & NT, F2F, FS, DF & Spatial & Single Frame & HRNet & NT, F2F,~FS, DF, GAN \\
 \textbf{FFD} & NT, F2F, FS, DF, GAN & Spatial & InsightFace, Single Frame & XceptionNet, VGG & NT, F2F,~FS, DF, GAN \\
\textbf{RECCE} & NT, F2F,~FS, DF & Spatial & RetinaFace, Single Frame & XceptionNet & NT, F2F,~FS, DF, GAN \\
 \textbf{CORE} & NT, F2F,~FS, DF, GAN & Spatial & MTCNN, Single Frame & XceptionNet & NT, F2F,~FS, DF, GAN \\
\textbf{IID} & NT, F2F, FS, DF & Spatial & RetinaFace, Single Frame & ResNet & NT, F2F,~FS, DF, FaSh, GAN \\
 \textbf{MCX-API} & NT, F2F, FOM, FS, DF & Spatial & MTCNN, Single Frame & XceptionNet & NT, F2F, FOM, FS, DF, FSGAN, GAN \\ 
\hline
\textbf{EffB4Att} & NT, F2F, FS, DF, GAN & Spatial & BlazeFace, Single Frame & EfficientNet, Siamese & NT, F2F,~FS, DF, GAN \\
 \textbf{LTW} & NT, F2F, FS, DF & Spatial & MTCNN, Single Frame & EfficientNet & NT, F2F,~FS, DF + GAN \\
\textbf{MAT} & NT, F2F, FS, DF & Spatial & RetinaFace, Single Frame & EfficientNet & NT, F2F,~FS, DF + GAN \\
 \textbf{DCL} & NT, F2F, FS, DF & Spatial & DSFD, Single Frame & EfficientNet & NT, F2F,~FS, DF + GAN \\
\textbf{SBIs} & FS & Spatial & \begin{tabular}[c]{@{}l@{}}dlib, RetinaFace,\\Single Face, Single Frame\end{tabular} & EfficientNet & NT, F2F,~FS, DF, FSGAN, GAN \\ 
\hline
 \textbf{MLAC} & NT, F2F, FS, DF & Spatial & dlib, Single Frame & XceptionNet, GAN learning & NT, F2F,~FS, DF \\
\textbf{FRDM} & NT, F2F, FS, DF & Spatial & dlib, Single Frame & \begin{tabular}[c]{@{}l@{}}XceptionNet, Dual Cross \\Modal Attention\end{tabular} & NT, F2F,~FS, DF, GAN,VAE \\
 \textbf{OST} & NT, F2F, FS, DF & Spatial & dlib, Single Frame & XceptionNet, Meta Training & NT, F2F,~FS, DF, GAN,VAE \\ 
\hline
\textbf{CADDM} & NT, F2F, FS, DF, FaSh & Spatial & MTCNN, Single Frame & ResNet, EfficientNet & NT, F2F,~FS, DF, FaSh, GAN \\
 \textbf{QAD} & NT, F2F, FS, DF, FaSh & Spatial & dlib, Single Frame & \begin{tabular}[c]{@{}>{}l@{}}ResNet, EfficientNet, \\Collaborative learning\end{tabular} & NT, F2F, FS, DF, FaSh, GAN \\ 
\hline
\textbf{ICT} & FS & Spatial & \begin{tabular}[c]{@{}l@{}}RetinaFace, Self-blend \\on real image, Single Frame\end{tabular} & Vision Transformer & NT, F2F, FS, DF, GAN,VAE \\
 \textbf{UIA-ViT} & NT, F2F, FS, DF & Spatial & dlib, Single Frame & Vision Transformer & NT, F2F,~FS, DF, GAN \\
\textbf{AUNet} & NT, F2F,~~FS, DF & Spatial & dlib, RetinaFace, Single Frame & Vision Transformer & NT, F2F,~FS, DF, FSGAN, GAN \\ 
\hline\hline
 \textbf{ADDNet-3d} & FS, DF, GAN & Spatial,~Temporal & MTCNN, Multiple Frames & Convolutional layers & FS, DF, GAN \\
\textbf{DeepRhythm} & NT, F2F, FS, DF, GAN & Spatial, Temporal & dlib, MTCNN, Multiple Frames & ResNet & NT, F2F,~FS, DF, GAN \\
 \textbf{S-IML-T} & NT, F2F,~FS, DF, GAN & Spatial, Temporal & dlib, MTCNN, Multiple Frames & XceptionNet & NT, F2F,~FS, DF, GAN \\
\textbf{TD-3DCNN} & NT, F2F,~FS, DF & Spatial, Temporal & MobileNet, Multiple Frames & 3D Inception & NT, F2F,~FS, DF, GAN \\
 \textbf{DIA} & NT, F2F,~FS, DF, GAN & Spatial, Temporal & \begin{tabular}[c]{@{}>{}l@{}}RetinaFace, 4 keypoints,\\Multiple Frames\end{tabular} & ResNet & NT, F2F,~FS, DF, GAN \\
\textbf{DIL} & NT, F2F,~FS, DF, GAN & Spatial, Temporal & dlib, MTCNN, Multiple Frames & ResNet & NT, F2F,~FS, DF,~ GAN \\
 \textbf{FInfer} & NT, F2F,~FS, DF & Spatial,~Temporal & dlib, Multiple Frames & Convolutional layers & NT, F2F,~FS, DF, GAN,VAE \\
\textbf{HCIL} & NT, F2F,~FS, DF & Spatial,~Temporal & \begin{tabular}[c]{@{}l@{}}dlib, MTCNN, Multiple Frames,\\Multiple Frames\end{tabular} & ResNet & NT, F2F,~FS, DF, GAN \\
 \textbf{AltFreezing} & NT, F2F, FS, DF & Spatial,~Temporal & \begin{tabular}[c]{@{}>{}l@{}}Temporal drop, Temporal repeat,\\Self-blend on real, Multiple Frames\end{tabular} & 3D ResNet & NT, F2F,~FS, DF, FaSh, VAE \\ 
\hline
\textbf{STIL} & NT, F2F, FS, DF, GAN & Spatial,~Temporal & MTCNN, Multiple Frames & \begin{tabular}[c]{@{}l@{}}ResNet, Spatial-Temporal\\Inconsistency\end{tabular} & NT, F2F,~FS, DF, GAN \\
 \textbf{LipForensics} & NT, F2F,~FS, DF, FaSh & Spatial,~Temporal & \begin{tabular}[c]{@{}>{}l@{}}RetinaFace, Face Alignment,\\Cropping Mouths, Multiple Frames\end{tabular} & \begin{tabular}[c]{@{}>{}l@{}}ResNet\\Temporal CNN\end{tabular} & NT, F2F,~FS, DF, FaSh, GAN, VAE \\
\textbf{FTCN} & NT, F2F,~FS, DF & Spatial,~Temporal & \begin{tabular}[c]{@{}l@{}}InsightFace, Face Alignment,\\Multiple Frames\end{tabular} & \begin{tabular}[c]{@{}l@{}}3D ResNet, \\Transformer Encoder\end{tabular} & NT, F2F,~FS, DF, FaSh,~ VAE \\ 
\hline
 \textbf{CCViT} & FS, DF,~GAN & Spatial & MTCNN, Single Frame & EfficientNet,~Vision Transformer & NT, F2F,~FS, DF, FaSh, GAN \\
\textbf{CLRNet} & NT, F2F,~FS, DF, GAN & Spatial,~Temporal & MTCNN, Multiple Frames & 3D ResNet & NT, F2F,~FS, DF, GAN \\
 \textbf{LTTD} & NT, F2F,~FS, DF & Spatial,~Temporal & MTCNN, Multiple Frames & Vision Transformer & NT, F2F,~FS, DF, FaSh, GAN,VAE \\ 
\hline\hline
\textbf{F3-Net} & NT, F2F,~FS, DF & Frequency & F2F (RGB Tracking), Single Frame & XceptionNet & NT, F2F,~FS, DF \\
 \textbf{FDFL} & NT, F2F,~FS, DF & Spatial,~Frequency & \begin{tabular}[c]{@{}>{}l@{}}RetinaFace, DCT transform,\\Single Frame\end{tabular} & XceptionNet & NT, F2F,~FS, DF \\
\textbf{ADD} & NT, F2F,~FS, DF, FaSh & Spatial,~Frequency & dlib, Single Frame & ResNet,~Knowledge Distillation & NT, F2F,~FS, DF, FaSh \\
 \textbf{LRL} & NT, F2F,~FS, DF & Spatial, Frequency & Single Frame & Convolutional layers & NT, F2F,~FS, DF + GAN \\ 
\hline
\textbf{TRN} & NT, F2F,~FS, DF & \begin{tabular}[c]{@{}l@{}}Spatial,~Temporal,\\Frequency\end{tabular} & \begin{tabular}[c]{@{}l@{}}dlib,\\Multiple Frames\end{tabular} & DenseNet, BiLSTM & NT, F2F,~FS, DF + GAN \\
 \textbf{SPSL} & NT, F2F,~FS, DF & Frequency & IDCT Transform, Single Frame & XceptionNet & NT, F2F,~FS, DF \\ 
\hline
\textbf{CD-Net} & NT, F2F,~FS, DF & \begin{tabular}[c]{@{}l@{}}Spatial,~Temporal,\\Frequency\end{tabular} & \begin{tabular}[c]{@{}l@{}}DCT and IDCT Transform,\\Multiple Frames\end{tabular} & SlowFast & NT, F2F,~FS, DF, GAN,VAE \\
 \textbf{SFDG} & NT, F2F,~FS, DF & Spatial, Frequency & dlib, Single Frame & \begin{tabular}[c]{@{}>{}l@{}}Information Interaction layers,\\Graph CNN, U-Net, EfficientNet\end{tabular} & NT, F2F,~FS, DF, GAN,VAE \\ 
\hline\hline
\textbf{RFM} & NT, F2F,~FS, DF, GAN & \begin{tabular}[c]{@{}l@{}}Spatial, Forgery\\Attention Map\end{tabular} & \begin{tabular}[c]{@{}l@{}}Suspicious Forgeries Erasing,\\Single Frame\end{tabular} & XceptionNet & NT, F2F,~FS, DF, GAN \\
 \textbf{FD2Net} & NT, F2F,~FS, DF & Spatial, 3D & 3DDFA, Single Frame & XceptionNet & NT, F2F,~FS, DF, GAN \\
\textbf{SOLA} & NT, F2F,~FS, DF & \begin{tabular}[c]{@{}l@{}}Spatial, Frequency,\\Noise Traces\end{tabular} & \begin{tabular}[c]{@{}l@{}}RetinaFace, ASRM,\\Single Frame\end{tabular} & ResNet & NT, F2F,~FS, DF, FaSh \\
 \textbf{AVAD} & Real Video & \begin{tabular}[c]{@{}>{}l@{}}Spatial, Temporal,\\Voice Sync\end{tabular} & \begin{tabular}[c]{@{}>{}l@{}}S3FD, Face Alignment,\\Multiple Frames\end{tabular} & \begin{tabular}[c]{@{}>{}l@{}}3D ResNet, VGG, \\Transformer Encoder\end{tabular} & FOM,~FS, FSGAN, GAN \\
\textbf{LGrad} & GAN & Gradient & Pre-trained StyleGAN, Single Frame & ResNet & NT, F2F,~FS, DF, GAN \\ 
\hline
 \textbf{LRNet} & NT, F2F,~FS, DF & \begin{tabular}[c]{@{}>{}l@{}}Temporal,\\Landmarks\end{tabular} & \begin{tabular}[c]{@{}>{}l@{}}dlib, Openface, \\Multiple Frames\end{tabular} & LRNet & NT, F2F,~FS, DF, VAE \\
\textbf{NoiseDF} & NT, F2F,~FS, DF & Noise Traces & RIDNet, Single Frame & Siamese & NT, F2F,~FS, DF, GAN,VAE \\
\hline\hline
\end{tabular}
}
\end{table*}

To structurally overview the overall published deepfake detectors, we introduce a new categorization methodology in Table~\ref{app_tab:DetectorFurtherDetails}. All of the appropriate deepfake detection methods provide some mutual series of processes and components to build their deep learning network. Based on this knowledge, a deepfake detector could be easily broken down into several key components: \textit{Deepfakes for Training, Artifacts, Data Pre-processing, Model Training, and Model Validation}.

\section{Details of Detectors' Performance}\label{sec:appendix-white-box}
To enhance the reader's comprehension of our analysis of 16 selected deepfake detectors across White-box and Black-box datasets, we have delineated their performances in Table \ref{tb:performance}. This table encompasses six standard metrics: ACC, ACC@best, AUC, F1, Precision, and Recall. Here, ACC@best refers to the highest achievable ACC across various thresholds, and it is at these optimal thresholds that we calculate the corresponding F1, Precision, and Recall metrics. It is important to note that at the optimal threshold, some methods were unable to identify any fake videos within a dataset, which is reflected by a Recall of 0 and yields undefined (NaN) values for both Precision and F1 scores. 
\begin{table*}[!t]
\centering
\caption{Performance of selected detectors in 6 performance metrics on a Stabilized dataset and In-the-wild dataset.}
\resizebox{2.\columnwidth}{!}{
\begin{tabular}{l|cccccc||cccccc}
\hline
 & \textbf{ACC} & \textbf{ACC@best} & \textbf{AUC} & \textbf{F1} & \textbf{Precision} & \textbf{Recall} & \textbf{ACC} & \textbf{ACC@best} & \textbf{AUC} & \textbf{F1} & \textbf{Precision} & \textbf{Recall} \\ \hline
 & \multicolumn{6}{c||}{\textbf{Xceptionnet}} & \multicolumn{6}{c}{\textbf{Capsule Forensics}} \\ \hline
DeepFaceLab & 79.01 & 90.12 & 94.95 & 80.85 & 100.00 & 67.86 & 86.42 & 90.12 & 91.71 & 80.85 & 100.00 & 67.86 \\ 
Dfaker & 66.67 & 71.60 & 66.58 & 51.06 & 63.16 & 42.86 & 83.95 & 87.65 & 86.73 & 76.60 & 94.74 & 64.29 \\ 
Faceswap & 69.14 & 75.31 & 74.26 & 60.00 & 68.18 & 53.57 & 90.12 & 95.06 & 98.45 & 92.86 & 92.86 & 92.86 \\ 
FOM-Animation & 65.43 & 65.43 & 35.31 & INF & N/A & 0.00 & 67.90 & 69.14 & 46.63 & 19.35 & 100.00 & 10.71 \\ 
FOM-Faceswap & 65.43 & 65.43 & 35.65 & N/A & N/A & 0.00 & 70.37 & 70.37 & 55.93 & 25.00 & 100.00 & 14.29 \\ 
FSGAN & 65.43 & 66.67 & 63.88 & 44.90 & 52.38 & 39.29 & 70.37 & 77.78 & 73.79 & 52.63 & 100.00 & 35.71 \\ 
LightWeight & 67.90 & 72.84 & 73.11 & 54.17 & 65.00 & 46.43 & 90.12 & 93.83 & 98.25 & 88.00 & 100.00 & 78.57 \\ 
\textbf{Avg.} & 68.43$_{(4.88)}$ & 72.49$_{(8.69)}$ & 63.39$_{(21.51)}$ & 58.20$_{(13.79)}$ & 69.74$_{(17.93)}$ & 35.72$_{(26.08)}$ & 79.89$_{(9.95)}$ & 83.42$_{(10.90)}$ & 78.78$_{(20.72)}$ & 62.18$_{(30.20)}$ & 98.23$_{(3.07)}$ & 52.04$_{(32.07)}$ \\ \hline
\textbf{In-the-wild} & 27.06 & 72.94 & 39.45 & 84.35 & 72.94 & 100.00 & 30.27 & 72.94 & 49.78 & 84.26 & 73.19 & 99.28 \\ 
\hline \hline
& \multicolumn{6}{c||}{\textbf{FTCN}} & \multicolumn{6}{c}{\textbf{LRNet}} \\
\hline
DeepFaceLab & 88.89 & 93.83 & 97.71 & 90.91 & 92.59 & 89.29 & 58.02 & 65.43 & 54.78 & N/A & N/A & 0.00 \\ 
Dfaker & 88.89 & 93.83 & 97.71 & 91.23 & 89.66 & 92.86 & 65.43 & 69.14 & 69.88 & 63.77 & 53.66 & 78.57 \\ 
Faceswap & 91.36 & 95.06 & 98.52 & 92.86 & 92.86 & 92.86 & 64.20 & 70.37 & 71.39 & 50.00 & 60.00 & 42.86 \\ 
FOM-Animation & 91.36 & 97.53 & 99.66 & 96.55 & 93.33 & 100.00 & 58.02 & 65.43 & 59.10 & N/A & N/A & 0.00 \\ 
FOM-Faceswap & 91.36 & 96.30 & 99.19 & 94.92 & 90.32 & 100.00 & 58.02 & 65.43 & 58.89 & N/A & N/A & 0.00 \\ 
FSGAN & 90.12 & 95.06 & 97.51 & 93.10 & 90.00 & 96.43 & 64.20 & 65.43 & 63.11 & N/A & N/A & 0.00 \\ 
LightWeight & 91.36 & 95.06 & 98.32 & 92.86 & 92.86 & 92.86 & 64.20 & 71.60 & 72.98 & 69.33 & 55.32 & 92.86 \\ 
\textbf{Avg.} & 90.48$_{(1.17)}$ & 95.24$_{(1.32)}$ & 98.37$_{(0.81)}$ & 93.20$_{(1.98)}$ & 91.66$_{(1.59)}$ & 94.90$_{(4.05)}$ & 61.73$_{(3.49)}$ & 67.55$_{(2.73)}$ & 64.30$_{(7.13)}$ & 61.03$_{(9.95)}$ & 56.33$_{(3.29)}$ & 30.61$_{(40.97)}$ \\ \hline
\textbf{In-the-wild} & 41.35 & 72.89 & 58.30 & 84.21 & 72.96 & 99.56 & 39.13 & 76.95 & 48.48 & 86.97 & 76.95 & 100.00 \\ 

\hline \hline
& \multicolumn{6}{c||}{\textbf{MAT}} & \multicolumn{6}{c}{\textbf{CLRNet}} \\
\hline
DeepFaceLab & 55.56 & 92.59 & 97.17 & 89.29 & 89.29 & 89.29 & 74.12 & 76.47 & 81.68 & 82.46 & 79.66 & 85.45 \\ 
Dfaker & 55.56 & 87.65 & 91.37 & 82.54 & 74.29 & 92.86 & 74.12 & 74.71 & 77.65 & 81.86 & 76.38 & 88.18 \\ 
Faceswap & 55.56 & 91.36 & 96.16 & 85.71 & 85.71 & 85.71 & 78.24 & 80.59 & 85.32 & 85.46 & 82.91 & 88.18 \\ 
FOM-Animation & 54.32 & 85.19 & 88.54 & 76.00 & 86.36 & 67.86 & 61.18 & 64.71 & 61.85 & 78.57 & 64.71 & 100.00 \\ 
FOM-Faceswap & 54.32 & 86.42 & 88.88 & 79.25 & 84.00 & 75.00 & 64.71 & 65.88 & 66.59 & 75.42 & 70.63 & 80.91 \\ 
FSGAN & 54.32 & 79.01 & 83.42 & 72.13 & 66.67 & 78.57 & 71.18 & 71.18 & 71.95 & 78.03 & 76.99 & 79.09 \\ 
LightWeight & 55.56 & 91.36 & 95.62 & 83.02 & 88.00 & 78.57 & 77.06 & 79.41 & 83.67 & 84.72 & 81.51 & 88.18 \\ 
\textbf{Avg.} & 55.03$_{(0.66)}$ & 87.65$_{(4.73)}$ & 91.59$_{(5.03)}$ & 81.13$_{(5.83)}$ & 82.05$_{(8.37)}$ & 81.12$_{(8.68)}$ & 71.52$_{(6.36)}$ & 73.28$_{(6.27)}$ & 75.53$_{(8.98)}$ & 80.93$_{(3.71)}$ & 76.11$_{(6.44)}$ & 87.14$_{(6.77)}$ \\ \hline
\textbf{In-the-wild} & 71.26 & 73.73 & 68.93 & 84.40 & 74.46 & 97.40 & 58.32 & 72.87 & 55.25 & N/A & N/A & 0.00 \\ 

\hline \hline
& \multicolumn{6}{c||}{\textbf{SBIs}} & \multicolumn{6}{c}{\textbf{ICT}} \\ \hline
DeepFaceLab & 65.43 & 95.06 & 98.05 & 93.10 & 90.00 & 96.43 & - & 71.60 & 64.15 & 29.41 & 83.33 & 17.86 \\ 
Dfaker & 65.43 & 85.19 & 90.30 & 81.82 & 71.05 & 96.43 & - & 71.60 & 67.25 & 33.33 & 75.00 & 21.43 \\ 
Faceswap & 65.43 & 92.59 & 96.50 & 87.27 & 88.89 & 85.71 & - & 70.37 & 67.52 & 29.41 & 83.33 & 17.86 \\ 
FOM-Animation & 43.21 & 65.43 & 37.47 & N/A & N/A & 0.00 & - & 65.43 & 47.91 & N/A & N/A & 0.00 \\ 
FOM-Faceswap & 54.32 & 65.43 & 58.69 & N/A & N/A & 0.00 & - & 65.43 & 60.04 & N/A & N/A & 0.00 \\ 
FSGAN & 65.43 & 83.95 & 90.03 & 78.69 & 72.73 & 85.71 & - & 66.67 & 56.00 & 6.90 & 100.00 & 3.57 \\ 
LightWeight & 65.43 & 92.59 & 96.50 & 85.19 & 88.46 & 82.14 & - & 70.37 & 66.37 & 29.41 & 83.33 & 17.86 \\ 
\textbf{Avg.} & 60.67$_{(8.74)}$ & 82.89$_{(12.60)}$ & 81.08$_{(23.56)}$ & 85.21$_{(5.49)}$ & 82.23$_{(9.47)}$ & 63.77$_{(43.91)}$ & - & 68.78$_{(2.82)}$ & 60.33$_{(7.42)}$ & 25.69$_{(10.64)}$ & 85.00$_{(9.13)}$ & 11.23$_{(9.55)}$ \\ \hline
\textbf{In-the-wild} & 41.51 & 73.05 & 55.27 & 84.41 & 73.02 & 100.00 & - & 88.30 & 61.32 & N/A & N/A & 0.00 \\ 

\hline \hline
& \multicolumn{6}{c||}{\textbf{CADDM}} & \multicolumn{6}{c}{\textbf{MCX-API}} \\ \hline
DeepFaceLab & 71.60 & 97.53 & 99.66 & 96.30 & 100.00 & 92.86 & 77.78 & 97.53 & 99.73 & 96.3 & 100.00 & 92.86 \\ 
Dfaker & 71.60 & 93.83 & 98.65 & 90.57 & 96.00 & 85.71 & 77.78 & 87.65 & 94.74 & 82.14 & 82.14 & 82.14 \\ 
Faceswap & 71.06 & 95.06 & 99.12 & 93.1 & 90.00 & 96.43 & 77.78 & 92.59 & 98.05 & 88.89 & 92.31 & 85.71 \\ 
FOM-Animation & 46.91 & 65.43 & 36.66 & N/A & N/A & 0.00 & 62.96 & 69.14 & 62.53 & 56.14 & 55.17 & 57.14 \\ 
FOM-Faceswap & 60.49 & 71.60 & 61.32 & 30.30 & 100.00 & 17.86 & 59.26 & 67.90 & 59.03 & 50.00 & 54.17 & 46.43 \\ 
FSGAN & 88.89 & 91.44 & 83.02 & 88.00 & 78.57  & 65.43 & 58.02 & 65.43 & 60.38 & N/A & N/A & 0.00 \\ 
LightWeight & 71.60 & 95.06 & 99.06 & 92.86 & 92.86 & 92.86 & 77.78 & 92.59 & 97.37 & 88.89 & 92.31 & 85.71 \\ 
\textbf{Avg.} & 66.13$_{(9.40)}$ & 86.77$_{(12.87)}$ & 83.70$_{(24.92)}$ & 81.03$_{(25.25)}$ & 94.48$_{(5.06)}$ & 66.33$_{(39.98)}$ & 70.19$_{(9.58)}$ & 81.83$_{(13.76)}$ & 81.69$_{(19.77)}$ & 77.06$_{(19.21)}$ & 79.35$_{(19.94)}$ & 64.28$_{(33.06)}$ \\ \hline
\textbf{In-the-wild} & 44.36 & 73.00 & 61.53 & 84.38 & 72.98 & 100.00 &  44.62	&  72.94 & 52.35 & 84.35 & 72.94 & 100.00 \\

\hline \hline
& \multicolumn{6}{c||}{\textbf{AltFreezing}} & \multicolumn{6}{c}{\textbf{LipForensics}} \\ \hline
DeepFaceLab & 83.95 & 90.12 & 95.22 & 86.21 & 83.33 & 89.29 & 61.21 & 57.59 & 88.93 & 28.09 & 35.90 & 23.08 \\ 
Dfaker & 86.42 & 98.77 & 99.66 & 98.18 & 100.00 & 96.43 & 60.12 & 56.51 & 91.36 & 30.77 & 35.90 & 26.92 \\ 
Faceswap & 86.42 & 98.77 & 99.80 & 98.18 & 100.00 & 96.43 & 60.12 & 56.51 & 89.14 & 30.77 & 35.90 & 26.92 \\ 
FOM-Animation & 85.19 & 92.59 & 97.10 & 89.29 & 89.29 & 89.29 & 60.12 & 55.79 & 90.14 & 32.38 & 35.90 & 29.49 \\ 
FOM-Faceswap & 86.42 & 93.83 & 97.24 & 91.80 & 84.85 & 100.00 & 61.21 & 56.51 & 89.86 & 30.77 & 35.90 & 26.92 \\ 
FSGAN & 86.42 & 96.30 & 99.39 & 94.34 & 100.00 & 89.29 & 59.04 & 55.06 & 94.64 & 33.87 & 35.90 & 32.05 \\ 
LightWeight & 86.42 & 98.77 & 99.66 & 98.18 & 100.00 & 96.43 & 60.12 & 54.34 & 91.71 & 35.24 & 35.90 & 34.62 \\ 
\textbf{Avg.} & 85.89$_{(0.97)}$ & 95.59$_{(3.48)}$ & 98.30$_{(1.79)}$ & 93.74$_{(4.83)}$ & 93.92$_{(7.79)}$ & 93.88$_{(4.48)}$ & 60.28$_{(0.75)}$ & 56.04$_{(1.08)}$ & 90.83$_{(1.98)}$ & 31.70$_{(2.36)}$ & 35.90$_{(0.00)}$ & 28.57$_{(3.83)}$ \\ \hline
\textbf{In-the-wild} & 43.46 & 72.63 & 60.24 & 84.12 & 72.67 & 99.85 & 36.25 & 71.83 & 58.65 & 83.48 & 72.94 & 97.57 \\ 

\hline \hline
& \multicolumn{6}{c||}{\textbf{LGrad}} & \multicolumn{6}{c}{\textbf{EffB4Att}} \\ \hline
DeepFaceLab & 45.68 & 65.43 & 53.44 & 65.43 & 100 & 50.80 & 96.30 & 97.53 & 99.73 & 96.30 & 100.00 & 92.86 \\ 
Dfaker & 49.38 & 65.43 & 51.62 & 65.82 & 98.11 & 50.91 & 92.59 & 92.59 & 97.24 & 88.89 & 92.31 & 85.71 \\ 
Faceswap & 45.68 & 65.43 & 50.13 & 65.82 & 98.11 & 50.91 & 97.53 & 100.00 & 100.00 & 100.00 & 100.00 & 100.00 \\ 
FOM-Animation & 44.44 & 69.14 & 51.48 & 68.42 & 98.11 & 51.67 & 71.60 & 72.84 & 72.78 & 50.00 & 68.75 & 39.29 \\ 
FOM-Faceswap & 43.21 & 65.43 & 43.26 & 65.43 & 100.00 & 50.80 & 72.84 & 77.78 & 78.37 & 62.50 & 75.00 & 53.57 \\ 
FSGAN & 43.21 & 66.67 & 50.94 & 66.25 & 100.00 & 51.04 & 79.01 & 86.42 & 92.39 & 80.70 & 79.31 & 82.14 \\ 
LightWeight & 45.68 & 66.67 & 51.15 & 66.67 & 98.11 & 51.16 & 97.53 & 100.00 & 100.00 & 100.00 & 100.00 & 100.00 \\ 
\textbf{Avg.} & 45.33$_{(2.10)}$ & 66.31$_{(1.38)}$ & 50.29$_{(3.26)}$ & 66.26$_{(1.05)}$ & 98.92$_{(1.01)}$ & 51.04$_{(0.31)}$ & 86.77$_{(11.84)}$ & 89.59$_{(10.95)}$ & 91.50$_{(11.32)}$ & 82.63$_{(19.59)}$ & 87.91$_{(13.33)}$ & 79.08$_{(23.64)}$ \\ \hline
\textbf{In-the-wild} & 48.36 & 73.00 & 49.67 & 72.98 & 100.00 & 56.96 & 44.04 & 73.05 & 57.38	& 84.36	& 73.14	& 99.64\\ 
\hline \hline
& \multicolumn{6}{c||}{\textbf{CCViT}} & \multicolumn{6}{c}{\textbf{ADD}} \\ \hline

DeepFaceLab & 87.65 & 96.30 & 99.33 & 94.74 & 93.10 & 96.43 & 100.00 & 100.00 & 100.00 & 100.00 & 100.00 & 100.00 \\ 
Dfaker & 83.95 & 90.12 & 95.01 & 83.02 & 88.00 & 78.57 & 96.30 & 98.77 & 99.80 & 98.18 & 100.00 & 96.43 \\ 
Faceswap & 86.42 & 95.06 & 98.65 & 94.74 & 93.10 & 96.43 & 100.00 & 100.00 & 100.00 & 100.00 & 100.00 & 100.00 \\ 
FOM-Animation & 76.54 & 81.48 & 77.22 & 69.23 & 75.00 & 64.29 & 65.43 & 69.14 & 46.83 & 19.35 & 100.00 & 10.71 \\ 
FOM-Faceswap & 80.25 & 85.19 & 86.59 & 76.92 & 83.33 & 71.43 & 65.43 & 70.37 & 45.15 & 29.41 & 83.33 & 17.86 \\ 
FSGAN & 82.72 & 83.95 & 92.79 & 73.47 & 85.71 & 64.29 & 77.78 & 77.78 & 74.87 & 58.54 & 92.31 & 42.86 \\ 
LightWeight & 87.65 & 95.06 & 98.99 & 92.86 & 92.86 & 92.86 & 100.00 & 100.00 & 100.00 & 100.00 & 100.00 & 100.00 \\ 
\textbf{Avg.} & 83.59$_{(4.14)}$ & 89.59$_{(6.08)}$ & 92.65$_{(8.18)}$ & 83.57$_{(10.71)}$ & 87.30$_{(6.69)}$ & 80.61$_{(14.56)}$ & 86.42$_{(16.36)}$ & 88.01$_{(14.83)}$ & 80.95$_{(25.58)}$ & 72.21$_{(36.07)}$ & 96.52$_{(6.48)}$ & 66.84$_{(41.43)}$ \\ \hline
\textbf{In-the-wild} & 62.87 & 72.94 & 66.10 & 84.35 & 72.94 & 100.00 & 33.49 & 73.73 & 49.76 & 84.42 & 74.41 & 97.54 \\ 

\hline
\end{tabular}
}
\label{tb:performance}
\end{table*}

\section{Influential Factors FAQs}\label{app_sec:moreIFs}
\textbf{What are the tradeoffs between Spatial approaches and Frequency approaches?}
As discussed in Sec. \ref{sec:frequency}, frequency artifacts are mostly utilized to provide supplementary information to support artifacts from other detector methodologies like spatial or spatiotemporal.\\
\textbf{Do any detector methodologies favor precision over recall? Why might that be the case?} Interestingly, analysis of our white-box experiments in Table \ref{tb:performance} in Appendix \ref{sec:appendix-white-box} indicate that only the spatiotemporal artifact models showed balanced precision and recall, with the remaining spatial, frequency, and special artifact models all favoring precision over recall. Only one model is the exception to this rule, and that is the spacial artifact model MAT, which was a reasonably good all-round performer on the gray, white and black-box experiments. 

Reasons for this behavior may be due to the fact that spatiotemporal models, by analyzing both spatial and temporal data, are likely to generalize better across a variety of deepfake techniques (which is supported by our white-box results), contributing to their balanced performance. In contrast, models focusing on specific types of artifacts (spatial, frequency, or special) may be optimized to detect deepfakes that prominently feature these artifacts, leading to high precision but potentially at the expense of recall when the deepfakes lacking these features are missed.\\
\textbf{For what use cases would some techniques be favorable over others?}
As discussed in items (2) and (3) of Sec. \ref{sec:gray-box}, identity-based methods such as ICT (CF \#5) can be a good choice when the target demographic at model deployment time aligns with the demographic the detector was trained on (eg. trained on celebrity faces and deployed to help safeguard against deepfake attacks on celebrities). On the other hand, if your training dataset is comprised of a demographic different to the target demographic at model deployment time, identity exclusion approaches like that used by the CADDM (CF \#4) model could be a good choice. As discussed in remark (2) of Sec. \ref{sub:inpact_factors}, for security-critical applications where high recall in a detector is paramount (in addition to high F1 and AUC), the spatiotemporal models FTCN (CF \#7), AltFreezing (CF \#6), CLRNet (CF
\#8), or CCViT (CF \#8) are a good choice, as is the spatial artifact model MAT(CF \#2).

\end{document}